\documentclass[10pt,journal,compsoc]{IEEEtran}
\ifCLASSOPTIONcompsoc
  \usepackage[nocompress]{cite}
\else
  \usepackage{cite}
\fi

\usepackage{algorithmic}
\usepackage{array}
\usepackage{mdwmath}
\usepackage{mdwtab}
\usepackage{eqparbox}
\usepackage{bbding}
\usepackage{hyperref}

\usepackage{amsmath,amsfonts,bm}

\def\eqref#1{equation~\ref{#1}}

\def\1{\bm{1}}

\def\rx{{\textnormal{x}}}

\def\rmA{{\mathbf{A}}}

\def\rmF{{\mathbf{F}}}

\def\rmH{{\mathbf{H}}}

\def\rmK{{\mathbf{K}}}

\def\rmQ{{\mathbf{Q}}}

\def\rmV{{\mathbf{V}}}
\def\rmW{{\mathbf{W}}}
\def\rmX{{\mathbf{X}}}

\def\vb{{\bm{b}}}

\def\evepsilon{{\epsilon}}

\def\mV{{\bm{V}}}

\def\mX{{\bm{X}}}

\DeclareMathAlphabet{\mathsfit}{\encodingdefault}{\sfdefault}{m}{sl}
\SetMathAlphabet{\mathsfit}{bold}{\encodingdefault}{\sfdefault}{bx}{n}

\def\gG{{\mathcal{G}}}

\def\gR{{\mathcal{R}}}

\def\gV{{\mathcal{V}}}

\def\sR{{\mathbb{R}}}

\newcommand{\pdata}{p_{\rm{data}}}

\newcommand{\E}{\mathbb{E}}

\usepackage{xr}
\usepackage{amsmath}

\usepackage{amssymb}
\usepackage{pifont}

\usepackage{subfig}
\usepackage{comment}
\usepackage{soul}
\usepackage{algorithmic}
\usepackage{hyperref}
\usepackage{booktabs}
\usepackage{threeparttable}
\usepackage{mathrsfs}
\usepackage{amsfonts}
\usepackage[ruled]{algorithm2e}

\usepackage{multirow}
\usepackage{graphicx}
\usepackage{color}
\usepackage{balance}
\usepackage{caption}
\usepackage{adjustbox}
\usepackage{soul}
\setstcolor{red}
\usepackage{makecell}
\usepackage{array}
\usepackage{utfsym}
\usepackage{multirow}
\usepackage{enumitem}
\usepackage{ragged2e}

\usepackage{color}
\usepackage{tikz}
\usepackage[edges]{forest}
\definecolor{hidden-draw}{RGB}{0,0,0}
\definecolor{hidden-pink}{rgb}{0.98, 0.94, 0.75}
\definecolor{level0}{rgb}{0.67, 0.88, 0.69}
\definecolor{level1}{rgb}{0.98, 0.92, 0.84}
\definecolor{level2}{rgb}{0.8, 0.8, 1.0}
\definecolor{level3}{rgb}{1.0, 0.71, 0.76}
\usepackage{amsthm}
\theoremstyle{definition}
\newtheorem{definition}{Definition}[section]
\newtheorem{concept}{Concept}

\definecolor{lawngreen}{rgb}{0.49, 0.99, 0.0}
\definecolor{pink}{rgb}{1, 0, 0.5}
\definecolor{airforce}{rgb}{0.36, 0.54, 0.66}

\hypersetup{
    linkbordercolor=airforce,
    urlbordercolor=pink,
    citebordercolor=green,
}

\usepackage{tikz}

\begin{document}

\title{Foundation Models for Weather and Climate Data Understanding: A Comprehensive Survey}

\author{Shengchao Chen,~\IEEEmembership{Member,~IEEE,} Guodong Long, Jing Jiang, Dikai Liu,~\IEEEmembership{Senior Member,~IEEE,} and Chengqi Zhang,~\IEEEmembership{Senior Member,~IEEE}
\IEEEcompsocitemizethanks{\IEEEcompsocthanksitem Shengchao Chen, Guodong Long, Jing Jiang, and Chengqi Zhang are with the Australian Artificial Intelligence Institute, School of Computer Science, Faculty of Engineering and Information Technology, University of Technology Sydney, Sydney, NSW, 2007, Australia. (Email: pavelchen@ieee.org, guodong.long@uts.edu.au, jing.jiang@uts.edu.au, chengqi.zhang@uts.edu.au)\protect
\IEEEcompsocthanksitem Dikai Liu is with the Robotics Institute, University of Technology Sydney, Sydney, NSW, 2007, Australia. (Email: dikai.liu@uts.edu.au)
\IEEEcompsocthanksitem Corresponding Author: guodong.long@uts.edu.au (Guodong Long)
\IEEEcompsocthanksitem \small{Github-\url{https://github.com/shengchaochen82/Awesome-Large-Models-for-Weather-and-Climate}}.
}
\thanks{Survey Version Dec. 4, 2023.}}

\IEEEtitleabstractindextext{%
\begin{abstract}
\begin{justify}
As artificial intelligence (AI) continues to rapidly evolve, the realm of Earth and atmospheric sciences is increasingly adopting data-driven models, powered by progressive developments in deep learning (DL). Specifically, DL techniques are extensively utilized to decode the chaotic and nonlinear aspects of Earth systems, and to address climate challenges via understanding weather and climate data. Cutting-edge performance on specific tasks within narrower spatio-temporal scales has been achieved recently through DL. The rise of large models, specifically large language models (LLMs), has enabled fine-tuning processes that yield remarkable outcomes across various downstream tasks, thereby propelling the advancement of general AI. However, we are still navigating the initial stages of crafting general AI for weather and climate. In this survey, we offer an exhaustive, timely overview of state-of-the-art AI methodologies specifically engineered for weather and climate data, with a special focus on time series and text data. Our primary coverage encompasses four critical aspects: types of weather and climate data, principal model architectures, model scopes and applications, and datasets for weather and climate. Furthermore, in relation to the creation and application of foundation models for weather and climate data understanding, we delve into the field's prevailing challenges, offer crucial insights, and propose detailed avenues for future research. This comprehensive approach equips practitioners with the requisite knowledge to make substantial progress in this domain. Our survey encapsulates the most recent breakthroughs in research on large, data-driven models for weather and climate data understanding, emphasizing robust foundations, current advancements, practical applications, crucial resources, and prospective research opportunities.
\end{justify}
\end{abstract}

\begin{IEEEkeywords}
Foundation Models, Weather \& Climate Analysis, Deep Learning, Time Series, Spatio-Temporal data, Earth System.
\end{IEEEkeywords}}

\maketitle

\tableofcontents


\begin{table}[tbh]
    \centering
    \begin{tabular}{c|c}
    \toprule
        \textbf{Full Name} & \textbf{Abbreviation} \\
    \midrule
        Artificial Intelligence & AI \\
        Machine Learning & ML \\
        Deep Learning & DL \\
        General Circulation Models & GCMs \\
        Numerical Weather Prediction & NWP \\
        Natural Language Processing & NLP \\
        Computer Vision & CV \\
        Large Language Models & LLMs \\
        Vision-Language Models & VLMs \\
        Segement Anything Model & SAM \\
        Foundation Models & FMs \\
        Weather \& Climate Foundation Models & WFMs \\
        Task-Specific Models & TSM \\
        Few-/Zero Shot Learning & FSL/ZSL \\
        Convolutional Neural Networks & CNNs \\
        Recurrent Neural Networks & RNNs \\
        Generative Advarsial Netowrks & GANs \\
        Gated Recurrent Unit & GRU \\
        Long Short-Term Memory & LSTM \\
        Diffusion Models & DMs \\
        Graph Neural Networks & GNNs \\
        Spatio-Temporal Graph & STG \\
        Temporal Knowledge Graph & TKG \\
        Vision Transformer & ViT \\
        Fully-Connected Feed-Forward Network & FFN \\
        Adaptive Fourier Neural Operator & AFNO \\
        Adaptive Neuro-Fuzzy Inference System & ANFIS \\
        Masked Auto-Encoder & MAE \\
        Autoregressive & AR \\
        Evolutionary Multi-Objective Optimization & EMO \\
        El Nino-Southern Oscillation & ENSO \\
        Madden-Julian Oscillation & MJO \\
        Explainable AI & XAI \\
        Independent and Identically Distribution & IID \\
        Differential Privacy & DP \\
        Self-Supervised Learning & SSL \\
        Semi-Supervised Learning & SML \\
        Full-Supervised Learning & FLSL \\
        Federated Learning & FL \\
        \bottomrule
    \end{tabular}
    \caption{Full names and abbreviations of key nouns.}
    \label{tab:abbrev}
\end{table}
\section{Introduction}
\label{sec:introduction}

\begin{figure*}[tbh]
    \centering
    \includegraphics[width=.9\textwidth]{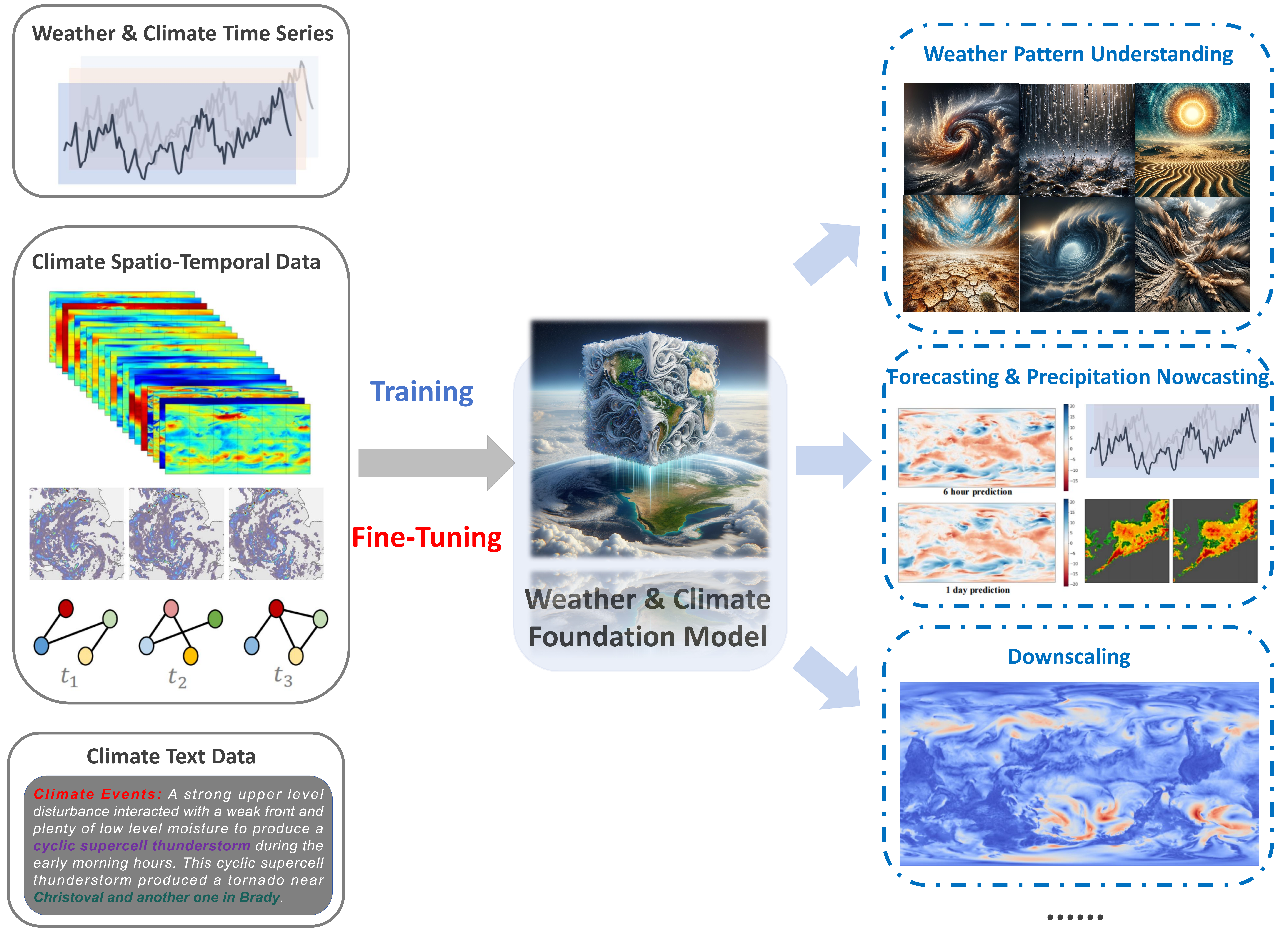}
    \caption{Large-scale Foundation Models for weather and climate data understanding can be both trained and skillfully redesigned and fine-tuned to handle a range of related downstream domains, such as forecasting, downscaling and weather pattern understanding, time series for text analysis, spatio-temporal data, and text data.}
    \label{fig:overall}
\end{figure*}

\begin{concept}
    \textit{Weather and Climate are distinct concepts with notable differences in spatial and temporal scales, variability, and predictability. The dissimilarities between the two can be elucidated as follows:}
    \begin{itemize}
        \item \textit{\textbf{Temporal Scale.} Weather pertains to the immediate state of atmospheric conditions, typically within a short-term timeframe. Conversely, climate represents a statistical summary of long-term weather patterns.}
        \item \textit{\textbf{Spatial Scale.} Weather represents atmospheric conditions at a specific location, whereas climate encompasses a comprehensive summary of typical weather patterns within a region over an extended period.}
        \item \textit{\textbf{Variability.} Weather exhibits rapid and frequent changes, while climate change occurs at a slower pace and encompasses long-term shifts in weather patterns.}
        \item \textit{\textbf{Predictability.} Weather prediction focuses on forecasting weather conditions in the next few days or shorter time scales. In contrast, climate prediction aims to forecast climate trends over the following months to decades.}
    \end{itemize}
\end{concept}

Climate change delineates noticeable alterations in global temperature and weather patterns over protracted periods. Currently, our planet is experiencing a proliferation in extreme natural phenomena, such as droughts~\cite{fabian2023modeling,deng2023divergent}, floods~\cite{fabian2023modeling}, earthquakes~\cite{zhou2023experimental}, heatwaves~\cite{barriopedro2023heat}, and intense rainfall~\cite{zeng2023response}, propelled by escalating climate change. Further amplifying these challenges are the alarming threats to ecosystems from mounting global warming and sea-level reductions~\cite{couldrey2023greenhouse,raihan2023review}. Given the projected augmentation in surface temperatures this century, we foresee an intensification in the harshness and frequency of these extreme phenomena~\cite{materia2023artificial}.

Leveraging advanced climate modeling and prediction techniques, which integrate a plethora of atmospheric and surface variables - encompassing atmospheric conditions, ocean currents, terrestrial ecosystems, and biosphere interactions - can enhance our comprehension of climate change~\cite{beddington2011achieving,connor2015united}. These insights can guide the formulation of bespoke mitigation strategies~\cite{kok2023mineral}. Long-term, accurate predictions of sea level changes can strengthen urban planning and disaster preparedness in coastal cities~\cite{loh2023analyzing,yu4575743grazing,voskamp2015planning}. In the short term, precise forecasts of rainfall, temperature, and humidity can heighten the safety of human activities, including agricultural planning and transportation scheduling~\cite{maier2000neural,markolf2019transportation,koetse2009impact}.

Traditionally, general circulation models (GCMs)~\cite{ravindra2019generalized} and numerical weather prediction models (NWPs)~\cite{bauer2015quiet,coiffier2011fundamentals,kimura2002numerical} have been favored tools for studying climate change trends and predicting future weather and climate scenarios. These models assimilate major Earth system components, including the atmosphere, surface, and oceans, to emulate the multidimensional dynamics of the Earth system. They identify potential nonlinear correlations between these components through complex physical equations, such as atmospheric dynamics, to generate predictions within a wide spectrum of physical parameters~\cite{krishnamurti2018introduction}. However, despite their considerable maturation, numerically constrained weather prediction models still encounter numerous challenges and limitations. One of these is their oversimplified representation of local geographical features~\cite{maraun2010precipitation}, as they often fail to capture the intricate nuances of local topography, which exerts a critical influence on regional weather and climate patterns. Another obstacle is the effective integration of observational data from disparate sources, such as weather stations, radars, and satellites~\cite{materia2023artificial}. Traditional models often struggle with incorporating these data, with varying spatial and temporal resolutions, into their modeling frameworks. Moreover, they require substantial computational resources to manage the myriad of physical constraints~\cite{chen2023prompt}. The complexity and scale of the Earth system demand extensive calculations, presenting challenges to computational capacity and efficiency.

The rapid advancement of AI technology has introduced cost-effective, direct, and simplified solution strategies for weather and cliamte modeling. In particular, Machine Learning (ML) and Deep Learning (DL) technologies can discern potential trend representations in weather and climate data, bypassing the need for intricate physical relationships.  Initially, ML techniques were sparingly used for short-term, localized forecasts of weather and climate conditions, given their limited capabilities compared with large-scale, time-extensive physical models. However, the past decade has witnessed an exponential surge in the application of data-driven deep learning methods in weather and climate research, propelled by the explosive expansion of global weather and climate data~\cite{nguyen2023climax,schultz2021can}. Capitalizing on abundant data resources and advancements in computational technology~\cite{wei2023licom3,prein2023towards}, these models are revolutionizing climate science~\cite{de2023review}. Employing voluminous data, deep learning models unravel the intricate nonlinear relationships concealed within climate variables, thereby capturing the dynamism and complexity of the climate system with enhanced precision~\cite{willard2020integrating,ren2021deep}. However, these models are often designed for specific tasks and trained with data in particular formats, such as regional weather forecasting or downscaling on a microscale. Differences in the representations of training data sources have resulted in an overly compartmentalized functionality of data-driven deep learning models for understanding weather and climate data. Consequently, it poses a significant challenge to develop a versatile climate model that can be fine-tuned for simulating the global weather and climate system.

The recent emergence and swift advancement of large models have yielded significant gains across various fields, including natural language processing (NLP), computer vision (CV)~\cite{yuan2021florence}, robotics~\cite{singh2023progprompt}, and a range of interdisciplinary areas encompassing life sciences~\cite{gilbert2023large,thirunavukarasu2023large,chen2023interpretable,zhang2023customized,ma2023segment}. Particularly in the NLP field, large models, or large language models (LLMs), are evolving rapidly, trained on large-scale corpora and fine-tuned for various downstream tasks~\cite{abburi2023generative,shi2023chatgraph,sun2023text}. In computer vision, large vision models trained on substantial natural images~\cite{lin2014microsoft,veit2016coco,deng2009imagenet} demonstrate exceptional zero-shot capabilities~\cite{kirillov2023segment,radford2021learning}. The impressive performance of these models across tasks arises from their substantial parameter counts and large-scale pre-training data. For instance, GPT-3~\cite{floridi2020gpt,brown2020language} possesses nearly 120 times the parameters of GPT-2~\cite{radford2019language}, enabling it to learn more powerfully from fewer samples, while GPT-4~\cite{bubeck2023sparks} has less than ten times the parameters of GPT-3, yet excels in text generation and image understanding. The rapid ascension of LLMs has redefined the path forward for deep learning, despite long-standing developments in areas such as unsupervised/semi-supervised and transfer learning. A notable example is the vision-language large model~\cite{zhu2023minigpt,radford2021learning,gao2022pyramidclip,zhang2021vinvl}, such as CLIP~\cite{radford2021learning}, which is trained on numerous natural image-text pairs and fine-tuned to achieve promising results in tasks like image segmentation~\cite{wang2022cris,park2022per,liang2023open} and video subtitle generation~\cite{tang2021clip4caption,zhang2022create}. Recently, the extension of large models into domains such as speech~\cite{ling2023adapting,zhang2023google}, physics~\cite{holmes2023evaluating}, and mathematical analysis~\cite{matzakos2023learning} has catalyzed advancements in fundamental science and specialized areas.

The groundbreaking success of pre-trained foundation models has propelled the domains of NLP and CV significantly closer to the realization of versatile AI. This advancement prompts an intriguing question:
The success of pre-trained foundation models has allowed the fields of NLP and CV to take a meaningful step towards realizing general-purpose AI, which not only leads one to wonder: \textbf{\textit{Is it possible to develop a universal foundation model for weather and climate data understanding that effectively addresses a myriad of related tasks?}} 

Building upon the theory of pre-trained models, \textsc{ClimaX}\cite{nguyen2023climax} introduces an innovative approach towards the development of a weather and climate base model. It leverages the Transformer to pre-train large-scale weather and climate data, yielding a flexible foundation model proficient in short- to medium-term forecasting,, climate projection, and downscaling. Both \textsc{PanGu-Weather}~\cite{bi2023accurate} and \textsc{W-MAE}~\cite{man2023w} exhibit robust climate prediction capabilities by modeling the global climate system using copious data. However, the quest for large-scale, universal climate models faces significant obstacles. A primary challenge is the scarcity of large, diverse, and high-quality training datasets. Existing datasets (refer to Table.~\ref{tab:dataset} for more details) struggle with inconsistent measurements, spatial-temporal biases, and limited functionality, hampering the progression of all-encompassing, multipurpose large-scale foundation models. Additionally, the computational demands of these models add another dimension of complexity, with the required infrastructure potentially unachievable in resource-limited settings. Ideally, a weather/climate foundation model should seamlessly handle multi-source observations and incorporate detailed representations of geographic features to generate more precise simulations of weather and climate trends. Unfortunately, this remains a largely uncharted territory for current weather and climate base models. Moreover, the interpretability of these models, often perceived as "black boxes," is a significant concern. In tasks related to weather and climate, where erroneous predictions can wreak havoc on ecosystems and societies, the need for interpretability is especially accentuated\cite{liu2023explainable,wang2023interpretable,chen2023interpretable}. Despite the remarkable strides and potential in understanding weather and climate data, the distinct challenges associated with the development of large-scale foundation models, as outlined above, necessitate concentrated research (refer to Sec.~\ref{sec:outlook} for more details). This emphasizes the need for a thorough review of advancements in this nascent field.

In this paper, we conduct a comprehensive review of data-driven models explicitly designed for weather and climate data. Our survey encompasses a wide array of large foundation models/task-specific models spanning various data types, model architectures, application domains, and representative tasks. This review amplifies the scope of insights derived from weather and climate data, encouraging novel strategies and fostering the cross-application of large models in the weather and climate. By leveraging the power of DL in large-scale models, we aim to reveal complex climate patterns, augment predictions, and deepen our comprehension of the climate system, thereby empowering society to more effectively adapt to the challenges posed by climate change. Our contributions are summarized as follows:
\begin{itemize}
    \item \textbf{First Comprehensive and Contemporary Survey.} To the best of our knowledge, this paper constitutes the inaugural comprehensive survey that thoroughly encapsulates the state-of-the-art developments of  large, and task-specific models for weather and climate data understanding, spanning across time series, video streams, and text sequences. We furnish an in-depth and current panorama that covers the broad spectrum of the domain, simultaneously delving into the subtleties of distinct methodologies, thereby providing the reader with a comprehensive and current apprehension of this field.


    \item \textbf{Systematic and In-depth Categorization.}  We introduce and discuss an organized and detailed categorization, dividing existing related research into two main categories: large climate foundation models and task-specific climate models. Furthermore, we further classify them based on the underlying model architectures, including RNNs, Transformers, GANs, Diffusion models, and Graph Neural Networks. Subsequent divisions are made based on the models' application domains and specific tasks, with detailed explanations of these task definitions. This multidimensional categorization provides readers with a coherent roadmap.
    
    \item \textbf{Abundant Resource Compilation.} e have assembled a substantial collection of datasets and open-source implementations pertinent to the field of weather and climate science. Each dataset is supplemented with an exhaustive description of its structure, pertinent tasks, and direct hyperlinks for expedient access. This compilation serves as an invaluable resource for prospective research and developmental endeavors in the domain.

    \item \textbf{Future Outlook and Research Opportunities.} We have delineated several promising trajectories for future exploration. These viewpoints span across various domains, including data post-processing, model architectures, interpretability, privacy, and training paradigms, among others.  This discourse equips the readers with an intricate understanding of the current status of the field and potential avenues for future exploration.

    \item \textbf{Insights for Designing.} We discuss and pinpoint crucial design elements for promising weather and climate foundation models. These design components incorporate the selection of temporal and spatial scales, dataset choice, data representation and model design, learning strategies, and evaluation schemes. Adherence to this systematic design pipeline enables practitioners to rapidly comprehend the design principles and construct robust weather and climate foundation models, thereby fostering the expeditious advancement of the weather and climate domain.
\end{itemize}

\begin{table*}[htbp]
  \centering
  \caption{Comparison between this and other related surveys, focusing on domains (i.e. specific vs. general), relevant data modalities (e.g., time series, graphs, video streams, and text), primary areas of focus (i.e., Weather and Climate Foundation Models (\textbf{WFM}) and Task-Specific Models (\textbf{TSM})), and available resources (i.e., dataset, and tools \& models).}
    \resizebox{1\textwidth}{!}{
    \begin{tabular}{cc|cc|cccc|cc|cc}
    \toprule
    \multirow{2}[2]{*}{\textbf{Survey}} & \multirow{2}[2]{*}{\textbf{Year}} & \multicolumn{2}{c|}{\textbf{Domain}} & \multicolumn{4}{c|}{\textbf{Modality of Weather Data}} & \multicolumn{2}{c|}{\textbf{Focus}} & \multicolumn{2}{c}{\textbf{Resource}} \\
       &    & \textbf{Specific} & \textbf{General} & \textbf{Time Series} & \textbf{Graphs} & \textbf{Video Streams} & \textbf{Text} & \textbf{WFM} & \textbf{TSM} & \textbf{Dataset} & \textbf{Tools \& Models} \\
    \midrule
    Ren et al.~\cite{ren2021deep} & 2021 &  \textcolor{green}{\Checkmark}  &  \textcolor{red}{\XSolidBrush}  &  \textcolor{green}{\Checkmark}  &   \textcolor{red}{\XSolidBrush}  &    \textcolor{red}{\XSolidBrush} &   \textcolor{red}{\XSolidBrush}  &   \textcolor{red}{\XSolidBrush}  &  \textcolor{green}{\Checkmark}  &  \textcolor{red}{\XSolidBrush}  & \textcolor{red}{\XSolidBrush} \\
    Fang et al.~\cite{fang2021survey} & 2021 & \textcolor{green}{\Checkmark}   &  \textcolor{red}{\XSolidBrush}  &  \textcolor{green}{\Checkmark}  &  \textcolor{red}{\XSolidBrush}  &  \textcolor{green}{\Checkmark}  &   \textcolor{red}{\XSolidBrush} &  \textcolor{green}{\Checkmark}   &  \textcolor{red}{\XSolidBrush}  & \textcolor{red}{\XSolidBrush}   & \textcolor{red}{\XSolidBrush} \\
    Bochenek et al.~\cite{bochenek2022machine} & 2022 &  \textcolor{green}{\Checkmark}  &  \textcolor{red}{\XSolidBrush}  &  \textcolor{green}{\Checkmark}  &  \textcolor{red}{\XSolidBrush}  &  \textcolor{red}{\XSolidBrush}  & \textcolor{red}{\XSolidBrush}   &  \textcolor{red}{\XSolidBrush}  &  \textcolor{green}{\Checkmark}  &  \textcolor{red}{\XSolidBrush}  &  \textcolor{red}{\XSolidBrush}\\
    Jaseena et al.~\cite{jaseena2022deterministic} & 2022 &  \textcolor{green}{\Checkmark}  &  \textcolor{red}{\XSolidBrush}  &  \textcolor{green}{\Checkmark}  &  \textcolor{red}{\XSolidBrush}  & \textcolor{red}{\XSolidBrush}   &  \textcolor{red}{\XSolidBrush}  &  \textcolor{red}{\XSolidBrush}  &  \textcolor{green}{\Checkmark}  &  \textcolor{green}{\Checkmark}  & \textcolor{red}{\XSolidBrush} \\
    Chen et al.~\cite{chen2023machinereview} & 2023 &  \textcolor{green}{\Checkmark}  &  \textcolor{red}{\XSolidBrush}  &  \textcolor{green}{\Checkmark}  &  \textcolor{green}{\Checkmark}  &  \textcolor{red}{\XSolidBrush}  &   \textcolor{red}{\XSolidBrush} &  \textcolor{green}{\Checkmark}  &  \textcolor{red}{\XSolidBrush}  &  \textcolor{red}{\XSolidBrush}  &  \textcolor{red}{\XSolidBrush}\\
    Jones et al.~\cite{jones2023ai} & 2023 &  \textcolor{green}{\Checkmark}  &  \textcolor{red}{\XSolidBrush}  &   \textcolor{green}{\Checkmark}  &  \textcolor{red}{\XSolidBrush}  &  \textcolor{red}{\XSolidBrush}  &  \textcolor{red}{\XSolidBrush}  &  \textcolor{red}{\XSolidBrush}  &  \textcolor{green}{\Checkmark}  &  \textcolor{red}{\XSolidBrush}  & \textcolor{red}{\XSolidBrush} \\
    Molina et al.~\cite{molina2023review} & 2023 &  \textcolor{green}{\Checkmark}  &  \textcolor{red}{\XSolidBrush}  &  \textcolor{green}{\Checkmark}  &  \textcolor{red}{\XSolidBrush}  &  \textcolor{red}{\XSolidBrush}  & \textcolor{red}{\XSolidBrush}   &  \textcolor{red}{\XSolidBrush}  &  \textcolor{green}{\Checkmark}  &  \textcolor{red}{\XSolidBrush}  &  \textcolor{red}{\XSolidBrush}\\
     Materia et al.~\cite{materia2023artificial} & 2023 &  \textcolor{green}{\Checkmark}   &  \textcolor{red}{\XSolidBrush}  &  \textcolor{green}{\Checkmark}  &  \textcolor{red}{\XSolidBrush}  &   \textcolor{red}{\XSolidBrush} &  \textcolor{red}{\XSolidBrush}  &  \textcolor{red}{\XSolidBrush}  & \textcolor{red}{\XSolidBrush}   &  \textcolor{red}{\XSolidBrush}  &  \textcolor{red}{\XSolidBrush}\\
     Mukkavilli et al.~\cite{mukkavilli2023ai} & 2023 &  \textcolor{red}{\XSolidBrush}  &  \textcolor{green}{\Checkmark}  &  \textcolor{green}{\Checkmark}  &  \textcolor{green}{\Checkmark}  &   \textcolor{red}{\XSolidBrush} &  \textcolor{red}{\XSolidBrush}  &  \textcolor{green}{\Checkmark}  & \textcolor{red}{\XSolidBrush}   &  \textcolor{red}{\XSolidBrush}  &  \textcolor{red}{\XSolidBrush}\\
    \midrule
    \textbf{Our Survey} & 2023 &  \textcolor{green}{\Checkmark}   &  \textcolor{green}{\Checkmark}  &  \textcolor{green}{\Checkmark}  &  \textcolor{green}{\Checkmark}  &  \textcolor{green}{\Checkmark}  &   \textcolor{green}{\Checkmark} &  \textcolor{green}{\Checkmark}  &  \textcolor{green}{\Checkmark}  &  \textcolor{green}{\Checkmark}  & \textcolor{green}{\Checkmark} \\
    \bottomrule
    \end{tabular}}
  \label{tab:related-work}%
\end{table*}

\textbf{Paper Organization.} The remainder of this survey is structured as follows: Section~\ref{sec:relatedwork} delineates the distinctions between our survey and other corresponding studies. Section~\ref{sec:background} instills the reader with fundamental knowledge on foundational models, primary depictions of weather and climate data, and related tasks. Section~\ref{sec:basemodel} expounds upon the core architecture of paramount models for weather and climate tasks. Section~\ref{sec:modelinfo}, we present a synopsis of the principal model classifications currently in use for weather and climate tasks, encompassing climate basic models and task-specific models. This section furnishes a holistic view of the field prior to probing into the complexities of individual methodologies. Section~\ref{sec:overview} imparts a concise introduction to climate basic models and task-specific models, further stratifying task-specific models based on dissimilar model architectures. Subsequently, Section~\ref{sec:applications} undertakes an extensive exploration of data-driven deep learning models for specific weather and climate tasks. Considering the lack of a unified and comprehensive index for weather and climate datasets, Section~\ref{sec:dataset} presents an exhaustive collection of dataset resources and introductions, aiming to impart convenience and efficiency for readers. Section~\ref{sec:outlook} delineates the challenges currently impeding the evolution of weather and climate basic models, as well as prospective future directions in this field. Section~\ref{sec:design} proposes a potential blueprint for the construction of weather and meteorological basic models, aiding contemplation and execution by practitioners, and fostering the development of climate foundation models. Finally, Sec.~\ref{sec:conclusion} provides a summary and concluding remarks on the content of the survey.

\section{Related Work and Differences}
\label{sec:relatedwork}
While numerous expansive surveys have been executed to model weather and climate-related data from various vantage points, none of them emphasize the broad-spectrum scope of weather data. For example, Ren et al.\cite{ren2021deep} undertook a survey on deep learning-based weather forecasting, focusing on neural network architecture design and spatial and temporal scales, yet it omitted models pertinent to the era of the weather data explosion. Both Fang et al.\cite{fang2021survey} and Jones et al.\cite{jones2023ai} reviewed deep learning-based weather forecasting within the confines of specific scenarios, namely extreme weather conditions and climate impacts on flood risk. Conversely, Bochenek et al.\cite{bochenek2022machine} and Jaseena et al.\cite{jacques2022deep} exclusively addressed and summarized machine learning/deep learning-based works concerning ordinary time series. Chen et al.\cite{chen2023machinereview} provided a survey of machine learning methodologies in weather and climate, but the focus remained restricted to forecasting tasks. Furthermore, Molina et al.\cite{molina2023review} primarily emphasized the application of machine learning in climate modeling, such as sources of predictability in climate variability models, feature detection, extreme weather and climate prediction, observational model integration, downscaling, and bias correction. Materia\cite{materia2023artificial} primarily centered on reviewing literature that employed machine learning techniques for extreme weather detection and understanding. These aforementioned surveys lack thorough investigation into the applications of foundational models in weather data understanding. Mukkavilli et al.\cite{mukkavilli2023ai} discussed the application of large models to weather and climate tasks and the architectural design, which bears similarity to our endeavor, but does not include more detailed task-specific models and a wider range of data modalities. Globally, these surveys also lack a structured delineation and an exhaustive discussion of deep learning-based models for weather data understanding, as well as adequate resources (datasets, open-source models and tools, etc.) that are either not provided or are limited in their availability. Given the recent multiplication of large-scale models in domains such as vision\cite{lee2020learning,kirillov2023segment}, audio~\cite{bubeck2023sparks}, and text~\cite{liang2023open}, our intention with this survey is to provide an exhaustive and up-to-date overview of large-scale models for weather data understanding, as well as a structured delineation, synthesis, and discussion of pertinent task-specific models, with the objective of establishing a robust foundation for the design of weather and climate base models. Our aim surpasses merely documenting recent advances; we also focus on available resources, practical applications, and potential research directions. Table.~\ref{tab:related-work} encapsulates the discrepancies between our survey and other analogous reviews.

\section{Background and Preliminary}
\label{sec:background}
This study aims to review the recent progress in implementing data-driven models, with a primary emphasis on DL techniques, to address weather and climate tasks. The objective is to illuminate potential pathways for developing foundation models dedicated to weather and climate data understanding. We direct our attention towards two principal categories of models in the weather and climate domains: large-scale foundational models and task-specific models. In this section, we commence by discussing these two model types and elucidate their distinctions and connections. Subsequently, we delineate weather and climate-related data types and representative tasks across diverse domains. We conclude with an introduction to four prevalent base model architectures employed in weather and climate tasks.


\subsection{Foundation Models}
Foundation Models (FMs) originated as pre-trained LLMs with a broad capability to undertake a myriad of downstream tasks through fine-tuning strategies. These models constitute a versatile class, separate from task-specific models, due to their capacity to accommodate a range of downstream tasks and integrate heterogeneous representations. The prowess of FMs can be classified into two categories: (1) Cross-Modal Representation and (2) Reasoning and Interaction.

\textbf{Cross-Modal Representation.} his category involves multi-modal models, including vision-language models (VLMs)\cite{zhu2023minigpt,zhou2022conditional,radford2021learning,maaz2023video}. These models merge and align linguistic and visual modalities, demonstrating a significant potential for modal unification. A prime example is CLIP (Contrastive Language-Image Pre-training)\cite{radford2021learning}, which concurrently trains on text and image data using the contrastive learning method. It displays substantial Zero-Shot Learning (ZSL) and Few-Shot Learning (FSL) abilities on downstream tasks. Another innovative model, SAM (Segment Anything Model)\cite{kirillov2023segment}, integrates the concept of prompting into visual tasks, yielding remarkable zero-shot segmentation performance. Models like InstructBLIP\cite{dai2023instructblip}, CoCa\cite{yu2022coca}, BEIT-3~\cite{wang2022image}, InstructGPT~\cite{ouyang2022training}, and LLaMa~\cite{touvron2023llama,touvron2023llama2} further expand the reach of cross-modal foundation models, accommodating a broader spectrum of tasks and modal representations. In weather prediction and climate change applications, data typically exhibit large-scale and multimodal characteristics, such as radar observations~\cite{chen2022dynamic, wu2021motionrnn}, satellite images~\cite{veillette2020sevir}, ground-based observatories~\cite{chen2023prompt,zhu2023weather2k}, and organized gridded data~\cite{rasp2020weatherbench,rasp2023weatherbench,nguyen2023climatelearn}. These characteristics provide impetus for the development of data-driven FMs for weather and climate tasks.

\textbf{Reasoning and Interaction.} FMs demonstrate exceptional reasoning and planning ablities, exemplified by models like CoT~\cite{wei2023chainofthought}, ToT~\cite{yao2023tree}, and GoT~\cite{besta2023graph}, in addition to task planning agents. This category also involves interaction abilities, encompassing operations and communication. This study emphasizes the application of data-driven FMs for weather and climate tasks. Nonetheless, this area remains  uncharted, offering abundant opportunities for innovation.

\subsection{Task-Specific Models}
Contrary to previously mentioned FMs, the majority of DL models for weather and climate are mainly domain-specific (e.g., global/regional precipitation forecasting, extreme weather comprehension, climate model downscaling). This survey classifies these task-specific models into two categories based on the nature of task for time series: (1) Time Series-based Weather and Climate Analysis; (2) Spatio-Temporal Series-based Weather and Climate Analysis. We also delineate an area for climate text data: Climate Text Analysis Tasks.

\textbf{Time Series-based Weather and Climate Analysis.} This category primarily comprises DL models for weather and climate analysis that leverage time series data. These models typically utilize weather time series data obtained from a single weather station to determine sequential relationships between one or multiple variables from past observations, thereby facilitating future trend predictions for specific weather variables.

A classic example of a data-driven model for weather forecasting is the Auto Regressive Integrated Moving Average (ARIMA)~\cite{zhang2003time}, which enables non-stationary data to become stationary through a differencing operation, and subsequently employs a combination of auto-regression and moving averages to model the time series. Given the significant seasonality often present in weather data, such as fluctuations in temperature and rainfall, Seasonal ARIMA (SARIMA)~\cite{chen2018time} and Seasonal ARIMA with eXogenous variables (SARIMAX)~\cite{chen2014daily} have been developed to model weather series, building upon seasonal auto-regression/moving average principles. Vector Autoregression (VAR) serves as an alternate method capable of modelling and predicting multiple correlated variables concurrently. Deep Learning-based models, such as families of Recurrent Neural Networks (RNNs)~\cite{shi2015convolutional,husken2003recurrent,hochreiter1997long}, convolutional neural network (CNN)-based architectures, and models based on the Transformer (e.g., Informer\cite{zhou2021informer}, Autoformer~\cite{chen2021autoformer}, Crossformer~\cite{zhang2022crossformer}, ETSFormer~\cite{woo2022etsformer}, Reformer~\cite{kitaev2020reformer}, FEDformer~\cite{zhou2022fedformer}), have exhibited superior performance when dealing with non-stationary time series. These models are particularly useful due to their lack of reliance on additional statistical knowledge and their efficiency in long-term forecasting.

\textbf{Spatio-Temporal Series-based Weather and Climate Analysis.} Another focal area is DL models for weather and climate analysis that employ spatio-temporal series. Unlike time-series data, spatio-temporal data covers weather variable observations across multiple locations over time, allowing for the extraction of intricate spatio-temporal patterns. In this context, continuous radar echoes or satellite images that represent independent weather times are also considered as spatio-temporal sequences.

Data-driven models designed for analysing spatio-temporal series for weather and climate analysis are often required to capture both temporal and spatial correlations. For instance, the convolutional LSTM~\cite{shi2015convolutional}, a variant of the LSTM, incorporates convolutional operations to the LSTM to capture additional spatial correlations. 3D Convolutional Neural Networks (3D-CNNs) are frequently employed to consider spatio-temporal correlations of sequences simultaneously. Spatio-Temporal Graph Neural Networks~\cite{yu2017spatio}, and other graph-based structures, effectively encode different spatial information into graphs that capture spatial correlations as well as temporal trends of weather variables. Transformer models utilize self-attention mechanisms to assess the importance of different locations and time points when making predictions~\cite{Chen_2023}. Recent advancements in the field have also seen the exploration of generative AI, such as generative adversarial networks~\cite{goodfellow2020generative} and diffusion models~\cite{dhariwal2021diffusion}, for weather prediction and climate change based on spatio-temporal sequences, owing to their excellent generative quality.

\subsection{Types of Weather and Climate Data}
Investigations into weather and climate typically necessitate the exploration of both temporal and textual data. The primary objectives of these tasks involve discerning the relationships between historical weather patterns — often characterized by numerous meteorological variables — and future changes. This process also includes the extraction of specific features from textual sequences to aid detailed subsequent analysis. In these scenarios, our discussion mainly revolves around three primary data types: time series, spatio-temporal, and textual data. n the context of weather and climate analysis, time series can be broadly divided into two types: univariate and multivariate. A univariate time series might be represented by the daily mean temperature at a single observation point, while a multivariate time series may include daily precipitation and humidity data collected from the same observation point. Here, we first discuss the definition of univariate/multivariate time series. Formally, we follow the definitions of time series data in Ref.~\cite{jin2023large}, which we summarize below.

\begin{definition}[\bf Time Series Data] For a single point observation, a univariate time series sole weather variables (such as temperature) $\rx = \{x_1, x_2, x_3, ..., x_T\} \in \sR^T$ is a sequence of $T$ time step indexed in time order, where $x_t \in \sR$ is the variable value of the time series at time $t$. A multivariate time series including different climate variables (i.e., temperature, humidity, precipitation, etc.) $\rmX = \{\rx_1, \rx_2, \rx_3, ..., \rx_T\} \in \sR^{T \times D}$ is a sequence of $T$ time steps indexed in time order but with $D$ dimensions (variables), in which $\rx \in \sR^D$ denotes the values of the time series at time $t$ along $D$ channels.
\end{definition}

Global climate data are often represented as spatio-temporal series, i.e., chaotic correlations with both temporal (change trend) and spatial dimensions (geographic location). We define two distinct Spatio-Temporal Series: univerate spatio-temporal series and multivariate spatio-temporal series. They are both sequence of data points organized by both temporal and spatial dimensions.
\begin{definition}[\bf Spatio-Temporal Series] For univerate spatio-temporal series, follow Definition 2.1, there exist $N$ points on the Earth system, at each point there exits a time series $\rx = \{x_1, x_2, x_3, ..., x_T\} \in \sR^T$, where $x_t \in \sR$, the spatio-temporal series is formulated as $\mX_u = \{\rx_1, \rx_2, \rx_3,..., \rx_N\} \in \sR^{T \times N}$. Similarly, for multivariate spatio-temporal sequences, the series can be formulated as $\mX_{mu} = \{\rmX_1, \rmX_2, \rmX_3,..., \rmX_N\} \in \sR^{T \times D \times N}$, where $\rmX_N$ denote the multivariate time series at the $N$ space point.
\end{definition}
Notably that graph-based structure usually utilized to construct a spatio-temporal series, such as spatio-temporal graphs (STGs), temporal knowledge graphs (TKGs), video streams, and others. In this survey, we mainly focus above-mentioned classes, which are higly representative and align closely with the current spatio-temporal series-based weather forecasting and climate analysis tasks. And we follow the Ref. to define STGs and TKGs firstly, as follows.
\begin{definition}[\bf Spatio-Temporal Graphs]
    A spatio-temporal graph $\gG = \{\gG_1, \gG_2, \gG_3, ..., \gG_T\}$ denotes a sequence of T static graph snapshots (also named time steps) indexed in time order, in which $\gG_t = (\gV_t,\evepsilon_t)$ presents a snapshot at $t$-th time step; $\gV_t$ and $\evepsilon_t$ are sets of nodes and edges at time $t$. The adjacent matrix represents the correlation between nodes in the graph and node feature matrices are defined as $\rmA_t \in \sR^{N \times N}$ and $\rmX_t \in \sR^{N \times D}$, where $\rmA_t = \{a_{i, j}^t\}$ and $a_{i, j}^t \neq 0$ if there is an edge between node $i$ and $j$. In addition, $N = |\gV_t|$ is the number of nodes and $D$ is the dimension of node features.
\end{definition}

\begin{definition}[\bf Temporal Knowledge Graphs]
    Follow the definition of STGs, a temporal knowledge graph $\gG = \{\gG_1, \gG_2, \gG_3, ..., \gG_T\}$ is a sequence of $T$ knowledge graph snapshots indexed in time order, where $\gG_t = (\epsilon_tm \gR_t)$ is a snapshot consisting of the entity and relation sets at time $t$. Specifically, $\epsilon_t$ encapsulates both subject and object entities, and $\gR_t$ presents the set of relations between them. In a temporal knowledge graph, entities and relations may posses different features, denoted by $\rmX \in \sR^{|\epsilon_t| \times D_e}$ and $\rmX_t^r \in \sR^{|\gR| \times D_r}$, where $D_e$ and $D_r$ are feature dimensions.
\end{definition}

Spatio-temporal video streams belong to a species of spatio-temporal series, which are represented as regular spatial shapes and sequences organized in time order. In weather forecasting and climate analysis tasks, regional contiguous weather radar echoes and satellite images that symbolize specific climate events belong to this type, and we define spatio-temporal video streams based on the definition of spatio-temporal sequences as follows.
\begin{definition}[\bf Spatio-Temporal Video Streams]  Assume a spatio-temporal video streams $\mV = \{\rmF_1, \rmF_2, \rmF_3, ..., \rmF_T\}$ is a set of continue frames that cover $T$ time steps indexed in time order, where $\rmF_t$ denotes the $t$-th frame (or time step). Each frame is viewed as a matrix of pixels\footnote{we only consider the image mode withou any knowledge there.} can be formulated as $\rmF_t \in \sR^{C \times H \times W}$, where $C,H,W$ denote the channels, height, and width of the frame, respectively.
\end{definition}

\begin{definition}[\bf Text Sequence]
Let $S$ be a text sequence, where each element in the sequence represents a word or character. The text sequence can be represented as $S = \{x_1, x_2, \ldots, x_n\}$, where $x_i$ represents the $i$-th element in the sequence. The length of the text sequence, denoted as (N), can be defined as $N = |S|$, where $|\cdot|$ represents the cardinality or number of elements in the sequence. Furthermore, each element in the text sequence can be represented as a one-hot encoded vector, denoted as $X$. The one-hot encoded vector $X_i$ for the $i$-th element in the sequence is a binary vector of length $M$, where $M$ represents the total number of unique words or characters in the text corpus. The one-hot encoded vector $X_i$ has a value of 1 at the position corresponding to the index of the word or character in the vocabulary, and 0 elsewhere.
\end{definition}

\subsection{Mainstream Tasks for Weather and Climate}
Based on the above definitions, we will present representative weather and climate analysis tasks associated with the above data types and structures.
\begin{itemize}
    \item \textbf{Weather/Climate Time Series Tasks.} Time series analysis forms the bedrock of weather and climate studies.  Researchers frequently harness this methodology to extract meteorological trends from sequential data, projecting these tendencies onto multiple variable values across a specified temporal span for granular analysis. This overarching task encompasses three subtasks: \textit{Forecasting}, \textit{Classification}, and \textit{Imputation}. In the forecasting task, the primary goal in to precisely predict a specific variable for a designated future temporal window grounded on historical observation. This task can be bifurcated, based on the magnitude of the prediction window, into short-term forecasting (typically spanning several hours to a few days) and long-term forecasting (generally a week or beyond). Short-term weather forecasting is often employed in immediate weather prediction and urban planning, whereas long-term forecasting predominantly serves climate studies, agriculture, and energy sectors. Subsequently, the classification task is aimed at mapping distinct meteorological phenomena, such as drought intensities, based on a historical chronology of atmospheric observations. Finally, the imputation task is structured to fill missing values in the series. This task exploits potential information embedded in the series, accounting for data gaps that might emanate from sensor malfunctions or severe climate events, among other factors.
    \item \textbf{Graph Structure-based Tasks.} The mainstream task of graph structure-based for climate change is forecasting. We explore graph structure-based tasks in terms of both STGs and TKGs, as previously mentioned. STGs and TKGs is extensions for representing and reasoning about spatio-temporal information, fusing the relationships between time, space, and entities into a unified graph structure. Forecasting tasks aim to infer weather conditions at future spatio-temporal points based on historical observations and model predictions. Such tasks involve multiple variables, such as temperature, humidity, and barometric pressure, as well as temporal and spatial dimensions. The key challenges of spatio-temporal map prediction tasks are how to effectively capture and model spatio-temporal dependencies and how to cope with data uncertainty and missingness.
    \item  \textbf{Spatio-Temporal Video Streams Tasks.} Video data stands as a crucial asset in the examination of climate change and weather forecasting. In meteorological contexts, spatio-temporal video streams typically manifest as sequences of frames that depict weather fluctuations over a fixed period. These sequences may include regularly shaped radar images, satellite images, and other types of weather-related visual data. Therefore, the primary interest in spatio-temporal video stream data lies in prediction tasks—namely, the forecasting of future images based on a series of past consecutive frames. The quintessential task in this context involves the prediction of imminent rainfall based on radar echoes or the extrapolation of satellite imagery.
    \item  \textbf{Climate Text Tasks.} The analysis of climate textual data, or climate text analysis, aspires to distill significant patterns and insights. This process encapsulates several subtasks including \textit{Sentiment Analysis}, \textit{Topic Modeling}, \textit{Information Extraction}, and \textit{Trend Analysis}. Sentiment analysis endeavors to preceive the sentiment or perspectives encapsulated in cliamte text data (e.g., public perceptions of climate change). Topic modeling, conversely, strives to identify and classify the cardinal themes or subjects broached within climate texts, thereby fostering a comprehensive understanding of pivotal focus areas.Information extraction constitutes the extraction of specific details from climate texts, such as instances of extreme weather events or particulars of climate policy. Finally, trend analysis concentrates on pinpointing and examining trends within climate texts, aiding in the monitoring of shifts in public dialogue, scientific research, or policy discussions over time. Collectively, these tasks converge to a deeper discernment of climate issues. The insights harvested can enlighten decision-making mechanisms, policy development, and initiatives to amplify public cognizance.
\end{itemize}
Considering the aforementioned types of weather and climate data, we will now expound on a variety of tasks pertinent to weather and climate analysis. Note that we have omitted the explicit outline and definition of the Climate Text Analysis task due to its closely related subtasks, and instead adopted the aforementioned Climate Task as a proxy for the Climate Text Analysis definition. A succinct description of each task is as follows:
\begin{itemize}
    \item \textbf{Forecasting Tasks.} These tasks span from a few hours (nowcasting) to days and weeks (short- and medium-range forecasting). They may include regional forecasting for continental states, counties, or cities. Sub-seasonal to seasonal prediction involves forecasting weather between 2 weeks and 2 months in advance, bridging the gap between weather forecasts and seasonal climate predictions, which is imperative for disaster mitigation.
    
    \item  \textbf{Precipitation nowcasting tasks.} Precipitation Nowcasting is a weather forecasting technique designed to predict precipitation over the next few hours. Unlike traditional weather forecasting, it focuses on short-term changes in precipitation, usually predicted on time scales of minutes to hours. This task employs data from radar systems, satellites, weather observation facilities, and numerical models, combined with image processing techniques, to predict the distribution, intensity, and movement of precipitation over a brief future period via real-time monitoring and analysis of atmospheric clouds and precipitation systems. Therefore, we have isolated it from the general forecasting task.

    \item  \textbf{Downscaling tasks.} Given the coarse spatial resolution of global climate models, they can only offer general estimates of climate conditions at local or regional scales. Simulations often exhibit systematic biases that diverge from trends in observed data. Downscaling climate models aims to generate locally precise climate information from global climate projections by correlating this climate information to observed local climate conditions. This process enhances the data's spatial and temporal resolution, rendering it more suitable for local and regional analysis.
    
    \item  \textbf{Bias correction tasks.} Bias correction is vital in weather and climate applications. It aims to minimize or eliminate systematic biases in model outputs and observational data, which emerge due to uncertainties in weather models and measurement errors. In weather forecasting, bias correction enhances the accuracy of model predictions by adjusting variables such as temperature and precipitation to match actual observations. In climate research, bias correction is crucial for aligning climate model outputs with observational data, facilitating accurate analysis of climate change trends, evaluation of model performance, and reliable predictions of future climate changes. Various methods, including statistical, machine learning, and deep learning techniques, can be employed for bias correction, tailoring the approach based on the specific application and data characteristics. By minimizing or eliminating systematic biases, bias correction improves the quality and reliability of weather and climate data.
    
    \item  \textbf{Weather pattern understanding tasks.} This task strives to analyze weather data to comprehend the variations and trends in weather patterns and the climate system. It involves modeling and analyzing various elements of the weather system, such as pressure, temperature, humidity, wind speed, and wind direction, to disclose their relationships and interactions. The objective is to identify and interpret different weather patterns, such as cyclones, fronts, and high-pressure systems, and deduce their impacts on weather changes and extreme weather events. By gaining a deeper understanding of weather patterns, we can enhance our knowledge of weather forecasting and climate change, providing decision-makers and researchers with more accurate and comprehensive information about the weather system.
\end{itemize}

\section{Basic Structure for Weather \& Climate}
\label{sec:basemodel}
Considering the different types of data present in weather and climate tasks, we mainly consider the use of Convolutional Neural Networks (CNNs), Recurrent Neural Networks (RNNs), Graph Neural Networks (GNNs), Transformers, Generative Adversarial Networks (GANs), and Diffusion Models to mine complex correlations from these data. In this survey, we mainly focus on Recurrent Neural Networks, Transformers, Generative Adversarial Networks, Graph Neural Networks, and Diffusion Models. Considering the particular representations of weather and climate data, we focus on spatio-temporal graphical neural networks in our discussion of GNNs.

\subsection{Recurrent Neural Networks}
Recurrent Neural Networks~\cite{medsker2001recurrent} (RNNs) are a neural network architecture specialized in processing sequential data. In RNNs, information is passed on all the time, enabling the RNN to utilize previous information to influence subsequent outputs. RNNs are fundamental modules in deep learning and are widely used in language modeling~\cite{de2015survey,mikolov2011extensions,mikolov2012context}, time series analysis~\cite{husken2003recurrent,hewamalage2021recurrent,lazcano2023combined}, and many other sequence-related tasks. RNNs have also pioneered the use of deep learning techniques to deal with weather and climate modeling~\cite{shi2015convolutional}. The update rule for a general RNN can be expressed as:
\begin{equation}
    h_t = \sigma(W_h x_t + U_h h_{t-1} + b_h),
\end{equation}
where $h_t$ is the hidden state at $t$-th time step, $x_t$ is the input at $t$-th time step, $W_h$ and $U_h$ are the weight matrices, $b_h$ is the bias, and $\sigma$ is a nonlinear activation function such as tanh or ReLU.

However, ordinary RNNs often encounter the problems of gradient vanishing and gradient explosion in practice, making it difficult to handle long sequences. To solve this problem, some improved RNN structures have been proposed, such as Long Short-Term Memory~\cite{hochreiter1997long} (LSTM) and Gated Recurrent Unit~\cite{chung2014empirical} (GRU). ConvLSTM~\cite{shi2015convolutional} and ConvGRU~\cite{shi2015convolutional} are variants that introduce convolutional operations into LSTM and GRU, allowing them to process spatially structured data such as images or videos, they usually have utilized to process weather spatio-temporal series data such as radar echo or satellite image sequences. In these models, fully connected operations are replaced by convolutional operations. For example, the update rule of ConvLSTM can be expressed as:  
\begin{equation}
\begin{aligned}
    f_t &= \sigma(W_{xf} * x_t + W_{hf} * h_{t-1} + b_f) \\
    i_t &= \sigma(W_{xi} * x_t + W_{hi} * h_{t-1} + b_i) \\
    o_t &= \sigma(W_{xo} * x_t + W_{ho} * h_{t-1} + b_o) \\
    \tilde{C}_t &= \tanh(W_{xc} * x_t + W_{hc} * h_{t-1} + b_c) \\
    C_t &= f_t \circ C_{t-1} + i_t \circ \tilde{C}_t \\
    h_t &= o_t \circ \tanh(C_t)
\end{aligned}
\end{equation}
where $f_t$, $i_t$, $o_t$, and $\tilde{C}_t$ are forgetting gates, input gates, output gates, and candidate memory cells, respectively, $*$ denotes the convolution operation, and $\circ$ denotes the Hadamard product.
The ConvGRU update rules can be represented as follows:
\begin{equation}
    \begin{aligned}
        r_t &= \sigma(W_{xr} * x_t + W_{hr} * h_{t-1} + b_r) \\
        z_t &= \sigma(W_{xz} * x_t + W_{hz} * h_{t-1} + b_z) \\
        \tilde{h}_t &= \tanh(W_{xh} * x_t + r_t \circ (W_{hh} * h_{t-1}) + b_h) \\
        h_t &= (1 - z_t) \circ h_{t-1} + z_t \circ \tilde{h}_t
    \end{aligned}
\end{equation}
where $r_t$ and $z_t$ are the reset and update gates, respectively. These gating mechanisms allow ConvGRU to handle long time dependencies more efficiently. These formulas show that ConvGRU first computes the reset and update gates at each time step, then computes the candidate hidden state $\tilde{h}_t$, and finally computes the new hidden state $h_t$. The update gate $z_t$ plays a role in determining how many new candidate hidden states to use when computing new hidden states.

\subsection{Diffusion Models}
Diffusion Models (DMs)~\cite{ho2020denoising,song2020denoising} have achieved promising achievements in extensive applications across a range of fields including computer vision~\cite{dhariwal2021diffusion,saharia2022palette,rombach2022high,croitoru2023diffusion}, natural language processing~\cite{hertz2022prompt,blattmann2022retrieval,li2023diffusion}, due to their efficacy in emulating intricate, high-dimensional data distributions. DMs comprise a category of probabilistic generative models and the core of these lie the principles of diffusion process, which are stochastic procedures delineating the continuous stochastic motion of particles over time. At the core of these models lie the principles of diffusion processes, which are stochastic procedures delineating the continuous stochastic motion of particles over time. These processes model spatial or temporal diffusion wherein particles incline towards transitioning from zones of high concentration to those with lower densities, facilitating a gradual assimilation or blending of quantities. The principal concept involves conducting a sequence of diffusion steps, with each step updating the data's probability distribution. This is accomplished by incorporating Gaussian noise into the current data samples and iteratively refining them. The noise addition in each diffusion step perturbs the data points, and the iterative refinement guides these perturbed points to gradually converge to the target distribution. This iterative process is akin to a random walk in the data space, where the random perturbations, guided by the model, eventually lead to the generation of new data points following the target distribution. 

Mathematically, a diffusion model describes a Markov chain that begins with the data and ends with noise. Let's denote the data as x and the noise as z. The Markov chain has the following form:
\begin{equation}
x_t = \sqrt(1 - dt) * x_(t-1) + \sqrt(dt) * z_t
\end{equation}
where $z_t$ is sampled from a standard Gaussian distribution, $dt$ is a small time step and t is the current step. The goal of the diffusion model is to learn the reverse transition of this Markov chain, i.e., to generate data from noise. This is done by estimating the conditional distribution $p(x_(t-1) | x_t)$ and sampling from it. With enough steps, the chain will transform the noise $z$ into the data $x$.

\subsection{Transformers}
Transformer is a DL model and has become a key infrastructure for existing state-of-the-art (SOTA) large models applied to NLP and other sequence-to-sequence tasks (i.e., weather forecasting)~\cite{wen2022transformers}. The key to this is its ability to handle dependencies between any part of the input sequence and any part of the output sequence without having to rely on the order of the sequences as in RNNs~\cite{kalyan2021ammus}.

Vanilla Transformer utilizes an encoder-decoder architecture, where both the encoder and decoder are comprised of a series of stacked blocks. Each Transformer layer is composed of a self-attention layer and a fully-connected feed-forward network (FFN). Additionally, the decoder block incorporates an additional cross-attention layer on top of the self-attention layer to capture information from the encoder. To facilitate information flow and alleviate the vanishing gradient problem, residual connections~\cite{he2016deep} and layer normalization modules are implemented between each layer.

\textbf{Multi-Head Self-Attention.} At the heart of the Transformer achitecture lies in the self-attention mechanism. This mechanism plays a pivotal role in capturing relationships within an input sequence. It accomplishes this by calculating attention scores for each element in the sequence in relation to the other elements. These scores are then utilized to assign weights to the input sequence, resulting in the generation of a new weighted sequence. The formula for the self-attention mechanism is as follows:
\begin{equation}
    \rmH = \text{Attention}(\rmQ, \rmK, \rmV) = \text{softmax}\left(\frac{\rmQ \rmK^T}{\sqrt{d_k}}\right)\rmV,
\end{equation}
where the $d_k$ denotes the dimension of the key, $\rmQ \in \sR^{n \times d_k}$, $\rmK \in \sR^{m \times d_k}$, $\rmV \in \sR^{m \times d_v}$ are the query matrix, key matrix and value matrix respectively, which are linear transformations of the same input sequence $\rmX \in \sR^{n \times d}$ (or feature matrix from the previous layer) based on three weight matrices $\rmW_q \in \sR^{d \times d_k}$, $\rmW_k \in \sR^{d \times d_k}$, $\rmW_v \in \sR^{d \times d_v}$, as
\begin{equation}
    \rmQ = \rmX \rmW_q, \rmK = \rmX \rmW_k, \rmV = \rmX \rmW_v,
\end{equation}
The attention score is obtained by computing the dot product of the query martix and key matrix, then dividing by $\sqrt{d_k}$ for scaling, and finally normalizing by softmax. 

Transformer uses multi-head self-attention with multiple sets of $\rmQ^{(i)}, \rmK^{(i)}, \rmV^{(i)}$, each set corresponding to a distinct set of linear transformation matrix $\rmW_q^{(i)} \in \sR^{d \times d_k}$, $\rmW_k^{(i)} \in \sR^{d \times d_k}$, $\rmW_v^{(i)} \in \sR^{d \times d_h}$, where $d_h$ is set to $\frac{d_v}{h}$, $h$ is the number of heads. The final output of the multi-head self-attention is obtained by projecting the concatenation of a series of $\rmH_i$ into a new feature space with a new weight matrix $\rmW_{proj} \in \sR^{d_v \times d_{proj}}$, as follows:

\begin{equation}
    \begin{aligned}
        \rmH &= \text{Multi-Head Self-Attention}(\rmQ, \rmK, \rmV) \\
             & = \text{Concat}(\rmH_1, \rmH_2, ..., \rmH_h)\rmW_{proj}, \\
        \rmH_i & = \text{Attention}(\rmQ^{(i)}, \rmK^{(i)}, \rmV^{(i)}.
    \end{aligned}
\end{equation}
For decoder, there is an additional mask mechanism that prevents query vectors from attending to the future positions yet to be decoded. In addition, an extra cross-attention following the self-attention, where the $\rmQ$ is derived from the output of the previous layer in the decoder, and the $\rmK$ and $\rmV$ are transformed from the output of the last layer of the encoder. It is designed to avoid foreseeing the true label while considering information from the encoder when encoding.

\textbf{Fully-connected Feed-Forward Layer.} Fully-connected feed-forward Layers following the attention layer is consists of linear transformation and a non-linear activation function. Denote the input matrix $\rmX \in \sR^{n \times d_i}$, the output of the feed-forward layer is
\begin{equation}
    \rmF = \text{FFN}(\rmX) = \sigma (\rmW_1 \rmX + \vb_1) + \vb_2,
\end{equation}
where $\sigma(\cdot)$ presents the activation function, and $\rmW_1 \in \sR^{d_i \times d_m}$, $\vb_1 \in \sR^{d_m}$, $\rmW_2 \in \sR^{d_m \times d_o}$, $\vb_2 \in \sR^{d_o}$ are all learnable parameters. 

\textbf{Residual Connection and Normalization.} Following each attention layer and each feed-forward layer, residual connection and layer normalization are applied. They conduct to retaining information when the model is considerably deep and thus guarantees the model performance. Formally, given a neural layer $f(\cdot)$, the residual connection and normalization layer is defined as
\begin{equation}
    \text{Add \& Norm}(\rmX, f) = \text{LayerNorm}(\rmX + f(\rmX)).
\end{equation}

\textbf{Transformer Layer.} The design of the Transformer model enables parallel processing of the entire sequence, eliminating the need for sequential processing of elements as in RNNs. This parallel processing enhances its efficiency in handling long sequences. By utilizing a multi-layer self-attention mechanism, the Transformer model effectively captures long-distance dependencies in sequences, which is crucial for tasks involving translation, summarization, and other sequence-to-sequence operations.

\subsection{Generative Adversarial Networks}
Generative Adversarial Networks (GANs)~\cite{goodfellow2020generative} aim to train a generative model via adversarial processes, it have widely used to image generation~\cite{de2023review, xu2018attngan, zhang2022diversifying}, super-resolution~\cite{he2022gcfsr, park2023content}, style transferring~\cite{zheng2022p,bafti2023biogan}, and image-based weather forecasting~\cite{cheng2023highway}. The fundamental concept of GANs involves training two NNs adversarially: a Generator $G$ and a Discriminator $D$. The objective of the Generator $G$ is to learn the underlying data distribution and generate novel samples accordingly. The discriminator ($D$)'s objective is to differentiate between the samples generated by the generator and the real samples.During training, the generator aims to produce samples that can effectively fool the discriminator, while the discriminator strives to enhance its ability to differentiate between real and generated samples. This process can be regarded as a two-player zero-sum game, ultimately leading to an equilibrium where the discriminator cannot distinguish between the generator-generated samples and the real samples.

The objective function of GANs can be expressed as the following optimization problem:
\begin{equation}
\begin{aligned}
\min_G \max_D V(D, G) = & \E_{x \sim \pdata(x)}[\log D(x)]  \\
                        &+ \E_{z \sim p_{z}(z)}[\log(1 - D(G(z)))],
\end{aligned}
\end{equation}
where $x$ is a sample from the true data distribution $\pdata$, $z$ is a sample from some a prior noisy distribution $p_{z}$, $G(z)$ is the sample generated by the generator using the noisy sample $z$, and $D_x$ is the discriminator's estimate of whether the sample $x$ (either the true sample or the generated sample) is the true sample. Training of GANs typically involves alternately optimizing two of this objective function. First, the generator is fixed and the discriminator is optimized. Then, fix the discriminator and optimize the generator. This process is repeated until some equilibrium is reached, at which point the samples generated by the generator should be indistinguishable from the true samples by the discriminator.

\subsection{Spatio-Temporal Graph Neural Networks}
Spatio-Temporal Graph Neural Networks (STGNNs)~\cite{yu2017spatio} is a concept in machine learning that combines spatial and temporal information using graph structures. It is particularly useful for analyzing data with both spatial and temporal dependencies. In STGNN, the basic concept involves representing the data as a graph, where each node represents a spatial location and the edges capture the spatial connectivity. Additionally, each node also contains temporal information, representing the state of the variable at different time steps.

\textbf{Spatial Graph Structure.} Let $\gG = (\gV, E)$ be the graph representing the spatial connections, where V is the set of nodes representing spatial locations, and E is the set of edges representing the spatial relationships. Each node $v_i$ represents the feature vector $x_i$ of the corresponding location i.

\textbf{Temporal Information.} Let $X = x_i^t$ be the set of feature vectors for all locations at time t. Each feature vector $x_i^t$ represents the state of the variable at location i and time t.

\textbf{Spatio-temporal Graph Convolution.} STGNN incorporates both spatial and temporal information through graph convolution operations, which capture the relationships between variables at different locations and time steps. The spatio-temporal graph convolution can be represented as:

\begin{equation}
    h_i^{t+1} = f(\sum_{j \in N(i)} w_{ij} \cdot h_j^{t} + b_i^{t}).
\end{equation}
Here, $h_i^{t+1}$ represents the updated feature vector of node $i$ at time $t+1$, $N(i)$ denotes the set of neighbors of node $i$, capturing the spatial connections between locations, $w_{ij}$ represents the weight between node $i$ and its neighbor $j$, indicating the strength of their relationship, $h_j^{t}$ represents the feature vector of the neighboring node $j$ at time $t$. $b_i^{t}$ is a bias term for node $i$ at time $t$. $f(\cdot)$ represents an activation function, such as ReLU or Sigmoid, applied element-wise to the sum of weighted inputs. The spatio-temporal graph convolution operation combines the spatial connectivity and temporal dependencies to effectively capture the evolving patterns and relationships in the data.

\begin{table*}[htbp]
  \centering
  \caption{List of representative models under mainstream applications for weather and climate data. Each column represents, in turn, data type, model category, method, scope, specific task/domain, base model, institution, and year of publication, respectively, noting that the base model is dominated by the primary starter module. More details available in Section.~\ref{sec:applications}.}
  \resizebox{1\textwidth}{!}{
    \begin{tabular}{c|c|cccccc}
    \toprule
    \textbf{Data} & \textbf{Category} & \textbf{Method} & \textbf{Scope} & \textbf{Task/Domain} & \textbf{Base} & \textbf{Institute} & \textbf{Year} \\
    \midrule
    \multirow{70}[12]{*}{\rotatebox{90}{Time Series}} & \multirow{7}[2]{*}{Large Foundation Models} & FourCastNet~\cite{pathak2022fourcastnet} & Task-Specific & Forecasting & RNN & Nvidia & 2022 \\
       &    & GraphCast~\cite{lam2022graphcast} & Task-Specific & Forecasting & GNN & Google & 2022 \\
       &    & PanGu-Weather~\cite{bi2023accurate} & Task-Specific & Forecasting & Transformer & HUAWEI & 2023 \\
       &    & FengWu~\cite{chen2023fengwu} & Task-Specific & Forecasting & Transformer & SHAI & 2023 \\
       &    & FuXi~\cite{chen2023fuxi} & Task-Specific & Forecasting & Transformer & FuDan & 2023 \\
       &    & W-MAE~\cite{man2023w} & Task-Specific & Forecasting & Transformer & UESTC & 2023 \\
       &    & ClimaX~\cite{nguyen2023climax} & General & Forecasting/Downscaling & Transformer & UCLA & 2023 \\
\cline{2-8}       & \multirow{63}[10]{*}{Task-Specific Models} & PGnet & Task-Specific & Forecasting & GAN & CAAC & 2021 \\
       &    & Graphino~\cite{cachay2021world} & Task-Specific & Forecasting & GNN & TUD & 2021 \\
       &    & TemperatureGAN~\cite{you2021temperature} & Task-Specific & Forecasting & GAN & FuDan & 2021 \\
       &    & Keisler GNN~\cite{keisler2022forecasting} & Task-Specific & Forecasting & GNN & NONE & 2022 \\
       &    & GE-STDGN~\cite{ni2022ge} & Task-Specific & Forecasting & RNN, GNN & SEU & 2022 \\
       &    & HiSTGNN~\cite{ma2023histgnn} & Task-Specific & Forecasting & GNN & SWJTU & 2022 \\
       &    & DWFH~\cite{venkatachalam2023dwfh} & Task-Specific & Forecasting & RNN & UHK & 2023 \\
       &    & SWINRDW~\cite{chen2023swinrdm} & Task-Specific & Forecasting & RNN, Diffusion Models & Alibaba & 2023 \\
       &    & SWINVRNN~\cite{hu2023swinvrnn} & Task-Specific & Forecasting & RNN & Alibaba & 2023 \\
       &    & PoET~\cite{ben2023improving} & Task-Specific & Forecasting & Transformer & ECMWF & 2023 \\
       &    & OceanFourCastNet~\cite{bire2023oceanfourcast} & Task-Specific & Forecasting & Transformer & MIT & 2023 \\
       &    & SEEDS~\cite{li2023seeds} & Task-Specific & Forecasting & Diffusion Model & Google & 2023 \\
       &    & Dyfussion~\cite{cachay2023dyffusion} & Task-Specific & Forecasting & Diffusion Model & UCSD & 2023 \\
       &    & DITTO~\cite{ovadia2023ditto} & Task-Specific & Forecasting & Diffusion Model & TelAviv & 2023 \\
       &    & TeleViT~\cite{prapas2023televit} & Task-Specific & Forecasting & Transformer & NOA & 2023 \\
       &    & MetePFL~\cite{chen2023prompt} & Task-Specific & Forecasting & Transformer & UTS & 2023 \\
       &    & FedWing~\cite{chen2023spatial} & Task-Specific & Forecasting & Transformer & UTS & 2023 \\
       &    & FuXi-Extreme~\cite{zhong2023fuxiextreme} & Task-Specific & Forecasting & Transformer & FuDan & 2023 \\
\cline{3-8}       &    & Meshfreeflownet~\cite{esmaeilzadeh2020meshfreeflownet} & Task-Specific & Downscaling & CNN & UCB & 2020 \\
       &    & Cdanet~\cite{hammoud2022cdanet} & Task-Specific & Downscaling & CNN & KAUST & 2022 \\
       &    & GPCHC~\cite{harder2022generating} & Task-Specific & Downscaling & GAN & Mila & 2022 \\
       &    & Bayesian AIG-Transformer~\cite{gerges2022novel}  & Task-Specific & Downscaling & Transformer & UH & 2022 \\
       &    & Deepesd~\cite{bano2022downscaling}  & Task-Specific & Downscaling & CNN & IFCA & 2022 \\
       &    & Multi-Varibales HP~\cite{gonzálezabad2023multivariable}  & Task-Specific & Downscaling & CNN & IFCA & 2023 \\
       &    & PhysicsDL~\cite{harder2023physics} & Task-Specific & Downscaling & RNN & Mila & 2023 \\
       &    & Torchclim~\cite{fuchs2023torchclim} & Task-Specific & Downscaling & CNN & UNSW & 2023 \\
       &    & ResDiff~\cite{mardani2023generative} & Task-Specific & Downscaling & Diffusion Model & SDU & 2023 \\
\cline{3-8}       &    & MetNet~\cite{sonderby2020metnet} & Task-Specific & Precipitation Nowcasting & CNN & Google & 2020 \\
       &    & Nowformer~\cite{parknowformer} & Task-Specific & Precipitation Nowcasting & Transformer & KAIST & 2021 \\
       &    & MPL-GAN~\cite{liu2020mpl}  & Task-Specific & Precipitation Nowcasting & GAN & JCU & 2021 \\
       &    & CMGAT~\cite{peng2021cngat} & Task-Specific & Precipitation Nowcasting & GNN & NUDT & 2021 \\
       &    & GDE~\cite{asperti2023precipitation} & Task-Specific & Precipitation Nowcasting & Diffusion Model & UNIBO & 2021 \\
       &    & DMSF-GAN~\cite{chen2022dynamic}  & Task-Specific & Precipitation Nowcasting & GAN & GBAMWF & 2022 \\
       &    & PCT-CYCLEGAN~\cite{choi2023pct} & Task-Specific & Precipitation Nowcasting & GAN & KMA & 2022 \\
       &    & Rainformer~\cite{bai2022rainformer}  & Task-Specific & Precipitation Nowcasting & Transformer & ZUT & 2022 \\
       &    & EarthFormer~\cite{gao2022earthformer} & Task-Specific & Precipitation Nowcasting & Transformer & HKUTS & 2022 \\
       &    & PTCT~\cite{yang2021ptct}  & Task-Specific & Precipitation Nowcasting & Transformer & SYSU & 2022 \\
       &    & MM-RNN~\cite{ma2023mm}  & Task-Specific & Precipitation Nowcasting & RNN & HIT & 2023 \\
       &    & STIN~\cite{jin2023spatiotemporal} & Task-Specific & Precipitation Nowcasting & RNN & UCAS & 2023 \\
       &    & TempEE~\cite{Chen_2023} & Task-Specific & Precipitation Nowcasting & Transformer & GBAMWF & 2023 \\
       &    & Preformer~\cite{jin2023preformer}  & Task-Specific & Precipitation Nowcasting & Transformer & UCAS & 2023 \\
       &    & PreDiff~\cite{gao2023prediff} & Task-Specific & Precipitation Nowcasting & Diffusion Model & HKUTS & 2023 \\
       &    & MetNet-3 & Task-Specific & Precipitation Nowcasting & CNN & Google & 2023 \\
\cline{3-8}       &    & ENSOTR~\cite{ye2021transformer} & Task-Specific & Weather Pattern Understanding & Transformer & FJNU & 2021 \\
       &    & All-seanson CNN~\cite{ham2021unified} & Task-Specific & Weather Pattern Understanding & CNN & CNU & 2021 \\
       &    & CNN-GRU~\cite{ahmed2021hybrid} & Task-Specific & Weather Pattern Understanding & RNN & USQ & 2021 \\
       &    & ENSO-GTC~\cite{mu2022enso} & Task-Specific & Weather Pattern Understanding & GNN & TJU & 2022 \\
       &    & DLHF~\cite{jacques2022deep} & Task-Specific & Weather Pattern Understanding & CNN & UCB & 2022 \\
       &    & ARIMA-LSTM~\cite{xu2022application} & Task-Specific & Weather Pattern Understanding & RNN & NCUWE & 2022 \\
       &    & ENSO-ConvGRU~\cite{wang2023convolutional} & Task-Specific & Weather Pattern Understanding & RNN & UIUC & 2023 \\
       &    & DK-STN~\cite{li2023dk} & Task-Specific & Weather Pattern Understanding & GNN & JLU & 2023 \\
       &    & STIEF~\cite{wang2023interpretable} & Task-Specific & Weather Pattern Understanding & RNN & UCAS & 2023 \\
       &    & XDL-CN~\cite{liu2023explainable} & Task-Specific & Weather Pattern Understanding & CNN & NEU & 2023 \\    
\cline{3-8}       &    & CUNet~\cite{han2021deep} & Task-Specific & Bias Correction & CNN & OUC & 2021 \\
       &    & SVMBC~\cite{yoshikane2022bias} & Task-Specific & Bias Correction & CNN & UTokyo & 2022 \\
       &    & Birectional GRU~\cite{li2022numerical} & Task-Specific & Bias Correction & RNN & WHU & 2022 \\
       &    & Bi-LSTM~\cite{yang2022correcting} & Task-Specific & Bias Correction & RNN & FuDan & 2022 \\
       &    & LSTM-TCN~\cite{blanchard2022multiscale} & Task-Specific & Bias Correction & RNN & Verisk & 2022 \\
       &    & BLSTM-AM-GS~\cite{han2022short} & Task-Specific & Bias Correction & RNN & CUST & 2022 \\
       &    & SRDRN~\cite{wang2022deep}  & Task-Specific & Bias Correction & CNN & AU & 2022 \\
       &    & DIcorrector-remapper~\cite{ge2022dl}  & Task-Specific & Bias Correction & CNN & WUSTL & 2022 \\
       &    & UNIT+QM~\cite{fulton2023bias}  & Task-Specific & Bias Correction & GAN & UOE & 2023 \\
       &    & Weathergnn~\cite{wu2023weathergnn}  & Task-Specific & Bias Correction & GNN & Alibaba & 2023 \\
    \midrule
    \multirow{5}[2]{*}{\rotatebox{90}{Text Data}} & \multirow{5}[2]{*}{Large Foundation Models} & ClimateBert~\cite{webersinke2022climatebert} & General & Climate Text Analysis & Transformer & FAU & 2021 \\
       &    & CliMedBert~\cite{fard2022climedbert} & General & Climate Text Analysis & Transformer & UNMC & 2022 \\
       &    & OceanGPT~\cite{bi2023oceangpt} & General & Climate Text Analysis & Transformer & ZJU & 2023 \\
       &    & ClimateBert-NetZero~\cite{schimanski2023climatebertnetzero} & General & Climate Text Analysis & Transformer & UZH & 2023 \\
       &    & FClimateBert~\cite{garridomerchán2023finetuning} & General & Climate Text Analysis & Transformer & UPC & 2023 \\
    \bottomrule
    \end{tabular}}
  \label{tab:models}%
\end{table*}%

\section{Overview and Categorization}
\label{sec:overview}
In this section, we provide an overview and categorization of DL models for weather and climate. Our survey is structured along three main dimensions: data types, model architectures, and application domains. A detailed synopsis of the related works can be found in the Table.~\ref{tab:models}. Based on the scope of application, we primarily divide the existing literature into two main categories: \textit{Large Foundation Models} and \textit{Task-Specific Weather and Climate Models}. Considering the task generality of weather/climate foundation models, we discuss them at a high level without further subdivisions. For task-specific weather/climate models, we categorize them based on specific underlying architectures to facilitate readers in indexing and referencing specific works according to model architectures, including \textit{Recurrent Neural Networks}, \textit{Generative Adversarial Networks}, \textit{Transformers}, \textit{Diffusion Models}, and \textit{Graph Neural Networks}. Subsequently, at the application level, we divide the existing literature into two main categories based on specific data categories: \textit{Time Series for Weather and Climate}\footnote{The scope of time series data includes spatio-temporal series data and spatio-temporal video stream data.} and \textit{Text for Weather and Climate}.

In the first category, we further dissect the existing literature into six primary classes predicated on the domains of application: \textit{Forecasting}, \textit{Precipitation Nowcasting}, \textit{Downscaling}, \textit{Data Assimilation}, \textit{Bias Correction}, and \textit{Weather Pattern Understanding}. For the second category, we explore it as a general subject (\textit{Climate Text Analysis}), refraining from subdividing it into different subtasks. This is because these often originate from pre-trained LLMs, and the specific task characteristics are typically delineated based on downstream datasets rather than the model itself.

\section{Models for Weather \& Climate}
\label{sec:modelinfo}
In this section, we will delve into the advancements of Foundation Models and Task-Specific Models for weather and climate data understanding. A categorization of representative methods and detailed information can be found in Table.~\ref{tab:models}.

\begin{figure*}[tbh]
    \centering
    \includegraphics[width=1\textwidth]{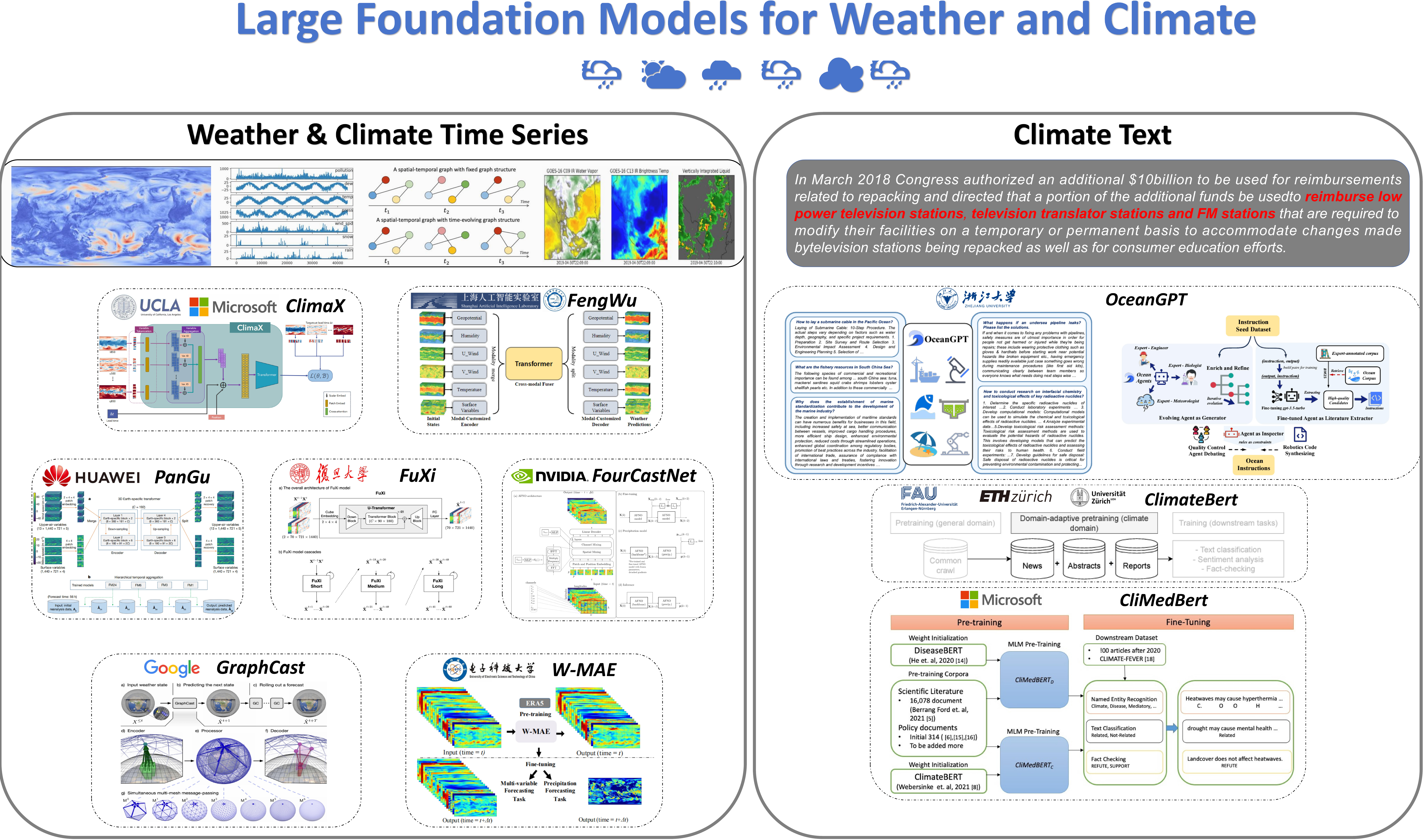}
    \caption{An overview of large foundation models for weather and climate, \textbf{Left:} Foundation Models specialized in weather and climate time series (including time series, spatio-temporal series, video streams, etc.), \textbf{Right:} Foundation Models specialized in climate-related text data.}
    \label{fig:fmsoverall}
\end{figure*}

\subsection{Foundation Models for Weather \& Climate}
The burgeoning development of foundation models in NLP~\cite{chen2021badpre,touvron2023llama,floridi2020gpt} and CV~\cite{kirillov2023segment,radford2021learning} has piqued research interest in foundation models for weather and climate data understanding. Large Foundation Models, created through pre-training strategies, can substantially enhance the generalization capability of AI-based climate models and can be fine-tuned for specific downstream tasks. Pre-training of such models necessitates large-scale sequence data, not typically sourced from ordinary time-series data.

Mindful of computational efficiency and the demand for timely climate predictions, Pathak et al. proposed \textsc{FourCastNet}\cite{pathak2022fourcastnet}, a climate pre-trained foundation model based on Vision Transformer and Adaptive Fourier Neural Network Operator (AFNO)\cite{guibas2021adaptive}, for high-resolution predictions and rapid inference. Its training process consists of self-supervised pre-training and autoregressive fine-tuning based on the pre-trained model. \textsc{Pangu-Weather}~\cite{bi2023accurate}, a data-driven model leveraging the 3D Earth-specific Transformer, is notable for its swift, precise global predictions and superior performance. It predicts atmospheric states over time based on the current state, described by five upper-air variables and four surface variables on a 0.25° horizontal grid with 13 vertical layers for the upper-air variables. On the other hand, \textsc{ClimaX}~\cite{nguyen2023climax} introduces the concept of fundamental modeling to weather prediction with its fully supervised pre-training based on the Transformer. It proposes variable disambiguation and variable aggregation strategies for merging and revealing potential relationships between different weather variations at various altitudes, offering promising flexibility for adapting to diverse downstream tasks, including global/regional/seasonal forecasting, climate mapping, and downscaling tasks. \textsc{FengWu}~\cite{chen2023fengwu} tackles the medium-term forecasting problem from a multimodal, multitask perspective with a uniquely designed deep learning architecture. It features a model-specific decoder and a cross-modal fusion Transformer that balances the optimization of different predictors in a regionally adaptive manner under the supervision of uncertainty loss. Given that the aforementioned large-scale models are trained via a fully supervised approach, \textsc{W-MAE}\cite{man2023w} implements unsupervised training of weather prediction models using a Masked Auto-Encoder (MAE)-based\cite{he2022masked,feichtenhofer2022masked} approach, which can be fine-tuned for downstream tasks through various data sources. MetePFL~\cite{chen2023prompt} and FedWing~\cite{chen2023spatial} also propose a Prompt-based federated learning~\cite{mcmahan2017communication} for training large foundation models, considerably reducing the cost of collaborative model training across regions while safeguarding data privacy. The rapid advancement of LLMs has led to the processing of weather and climate tasks that are no longer restricted to visual or time-series models. \textsc{OceanGPT}\cite{bi2023oceangpt}, based on LLMs, proposes a methodology for processing a wide range of ocean-related tasks. Beyond the foundation models used for forecasting and simulation, \textsc{ClimateBert}\cite{webersinke2022climatebert} is an NLP-based foundation model for processing climate-related texts. It is trained on over 2 million climate-related paragraphs from diverse sources such as news articles, research papers, and company climate reports~\cite{varini2021climatext}.

\subsection{Task-specific Models for Weather \& Climate}
In the realm of weather and climate analysis, task-specific models have been utilized for a myriad of specific tasks. This section will delve into the progress made in task-specific models for weather and climate, focusing on these principal architectures: RNNs, Transformers, GANs,  Diffusion Models, and Graph Neural Networks (GNNs).
\begin{itemize}
    \item \textbf{Recurrent Neural Networks (RNNs).} RNNs serve as the backbone of numerous weather forecasting models~\cite{diaconu2022understanding,tekin2021spatio,su2020convolutional,wang2022predrnn,wu2021motionrnn,wang2018eidetic,shi2015convolutional,luo2022predrann,jin2023spatiotemporal, venkatachalam2023dwfh,bilgili2022time,usharani2023ilf}. In addition to weather and climate prediction models built on RNNs architectures, hybrid models fusing RNN with other mechanisms have also gained traction~\cite{hu2023swinvrnn, chen2023swinrdm, Tang_2023_ICCV,zhifeng2022comparison,tian2019generative,leinonen2020stochastic}. For instance, the amalgamation of Swin Transformer~\cite{liu2021swin} with RNN has given birth to models like SwinVRNN~\cite{hu2023swinvrnn}, which capitalize on the advantages of both architectures. Moreover, the fusion of SwinRNN with generative models has led to models for the diffusion model SwinRDM~\cite{chen2023swinrdm} and for GAN~\cite{tian2019generative,leinonen2020stochastic}. Added to this, physical-informed based approaches have been introduced~\cite{zantedeschi2020towards}. Concurrently, with the evolution of Transformer-based spatio-temporal extraction, the integration of RNN architecture and Transformer models to address this problem has been on the rise~\cite{Tang_2023_ICCV, zhifeng2022comparison}.

    \item \textbf{Diffusion Models.} Standard diffusion models, comprising forward noisy processes and backward denoising processes, are widely employed for learning data distribution and generating data representations in meteorological and climatic contexts~\cite{leinonen2023latent,li2023seeds, cachay2021climart,lippe2023pde,gao2023prediff,ovadia2023ditto,chen2023swinrdm,hu2023swinvrnn,hatanaka2023diffusion}~\cite{mardani2023generative,hoivang2023diffmet}. For instance, SwinRDM~\cite{chen2023swinrdm} amalgamates SwinRNN~\cite{hu2023swinvrnn} and diffusion models to attain high-resolution weather forecasting. However, it is important to note that the application of diffusion models in weather and climate studies is still in its nascent stage. 

    \item \textbf{Generative Adversarial Networks (GANs).} GANs have widely used in image generation tasks, ranging from generating handwritten digits~\cite{radford2015unsupervised} to generating large-scale image datasets~\cite{brock2018large,karras2018progressive}. They are commonly employed in weather and climate tasks for spatiotemporal video stream prediction~\cite{bihlo2021generative,gupta2020climate}, aiming to generate realistic and temporally coherent sequences and match high-dimensional data distributions between them. Therefore, GAN-based architecture is common in weather and climate prediction tasks aims to generate predicted future frames like ground-truth as same as possible~\cite{klemmer2021generative,ravuri2021skilful, chen2022dynamic,klemmer2022spate} \cite{liu2020mpl}\cite{ji2023clgan,choi2023pct,luo2022experimental,wang2023skillful,tian2019generative,leinonen2020stochastic,harris2022generative,annau2023algorithmic} \cite{dai2022mstcgan,kim2021very}. Additional physical constraints are often introduced to improve the accuracy of weather and climate modeling in these hybrid models~\cite{hess2022physically,gupta2020climate,besombes2021producing,balogun2023temperaturegan,sleeman2023generative,meng2021physics,meng2023physical,lutjens2021physically,yuan2023space,chen2021physics}. 
    \item  \textbf{Transformers.} Transformer-based models are widely used for tasks related to time series analysis due to its powerful long series modeling capabilities, which also include responding to weather and climate change~\cite{bire2023oceanfourcast}. It focuses on short-term/long-term forecasting tasks in weather and climate applications and can be categorized into two types, the former focusing on one-/two-dimensional forecasts of weather and climate, such as predicting trends in relevant weather variables globally or regionally on single atmosphere level, and the latter focusing on multidimensional forecasts, such as extrapolations based on radar-echo imagery~\cite{lin2023comprehensive}, satellite cloud images~\cite{kuccuk2023transformer} and multi-layer atmosphere status, thus contributing to the understanding of weather patterns in the region. For the first category, the Transformer is used to perform short- and long-term forecasts, modeling dependencies on variables at different points in time through positional coding as well as self-attention mechanisms~\cite{bojesomo2021spatiotemporal,chattopadhyay2021physically,bilgin2021tent,ye2021transformer,bojesomo2023novel,gao2023spatio}. As for the second category, Transformers are expected to establish complex multilayered spatio-temporal relationships of meteorological variables at different atmospheric pressures, and the results of this type of Transformer are usually challenged based on the characteristics of the data itself (atmospheric pressures, spatio-temporal correlations, variable correlations), and so on~\cite{bi2023accurate,nguyen2023climax,chen2023fengwu,man2023w,ben2023improving}. Inspired by the fields of NLP and CV, the Transformer structure has also been redesigned for the development of large-scale weather and climate foundation models~\cite{nguyen2023climax,bi2023accurate,chen2023fengwu}. In addition, in the filed of NLP-based climate text analysis, Transformers is a general architecture~\cite{fard2022climedbert,vaghefi2023chatclimate,schimanski2023climatebertnetzero,krishnan2023climatenlp,fard2022climedbert,garridomerchán2023finetuning, auzepy2023evaluating, kraus2023enhancing}.
    
    \item \textbf{Graph Neural Networks.} In the field of weather and climate, numerous studies have explored the application of graph neural networks, particularly spatial-temporal graph neural networks, due to their ability to establish potential spatial-temporal relationships of the Earth system.~\cite{mu2022enso}. Two common applications include spatial-temporal sequence prediction~\cite{keisler2022forecasting, ni2022ge,wilson2018low,ma2023histgnn, ayadi2022wekg,li2023regional, wang2023convolutional, oskarsson2023graph,li2023dk,han2021joint,han2022semi,lam2022graphcast,cachay2021climart} and spatial-temporal video stream prediction in weather forecasting~\cite{peng2021cngat}. In spatial-temporal sequence prediction, graph neural networks are used to model the spatio-temporal dependencies and correlations in weather data. This involves predicting future weather conditions based on historical observations at different locations~\cite{chen2023prompt,chen2023spatial}. The graph structure is used to capture the spatial relationships between nodes, and the temporal dependencies are modeled using recurrent~\cite{han2021joint,han2022semi} or convolutional layers~\cite{ma2023histgnn,wang2023convolutional}. In spatial-temporal video stream prediction, graph neural networks are employed to predict future weather conditions in the form of video-like sequences~\cite{peng2021cngat}. This involves predicting the evolution of weather patterns over time, taking into account both spatial and temporal dependencies. 
\end{itemize}

\section{Applications}
\label{sec:applications}
This section presents an overview of prevalent DL models, categorized by their applications in weather and climate analysis. These applications include forecasting, precipitation nowcasting, downscaling, bias correction, data assimilation, climate text analysis, and weather pattern understanding.

\subsection{Forecasting}
Accurate weather and climate forecasting is critical for environmental and societal planning.  Significant strides have been made in developing robust DL methods that model the nonlinear associations between historical and future weather patterns. This section mainly focuses on discuss the advancement in the task of weather and climate forecasting based on time series and spatio-temporal series. The most common in such tasks are RNNs-based architecture, which are widely used due to their autoregressive (AR) architecture~\cite{chen2023swinrdm, venkatachalam2023dwfh,bilgili2022time,usharani2023ilf,hu2023swinvrnn}. For instance, DWFH introduces conductive long and short-term memory models to enhance data-driven deep weather prediction models~\cite{venkatachalam2023dwfh}. Ref.~\cite{bilgili2022time} merges the LSTM and an adaptive neuro-fuzzy inference system (ANFIS) for atmospheric pressure forecasting. SwinRDM introduces the SwinRNN as a fundamental component for high-resolution weather forecasting~\cite{chen2023swinrdm}, and diffusion models to achieve high-resolution weather forecasting at 0.25 degrees using a two-step training strategy: first, cyclic prediction of future atmospheric fields is performed at low resolution, followed by high-resolution and fine-grained atmospheric detail reconstruction based on the diffusion-based super-resolution model. Moreover, \textsc{Swinvrnn} employs a Recurrent Neural Network-based architecture with variations loss to improve long-lead weather forecasts~\cite{hu2023swinvrnn}. In addition, Transformer, especially Vision Transformer, is also widely used in weather and climate prediction based on spatio-temporal series due to its bright performance in modeling potential representational associations between image regions using Patch mechanism and self-attention mechanism. \textsc{FourCastNet}~\cite{pathak2022fourcastnet} delivers impressive performance in various weather forecasting tasks using 0.25° resolution. This achievement is based on the Vision Transformer (ViT)\cite{dosovitskiy2020image} and Adaptive Fourier Neural Network Operators (AFNO). \textsc{PoET}~\cite{ben2023improving} introduces hierarchical ensemble transformers to enhance medium-range ensemble weather forecasts on a global scale. \textsc{TeleViT}~\cite{prapas2023televit} integrates fine-grained local-scale and global-scale inputs, treating the Earth as one interconnected system for seasonal wildfire forecasting. Large models came out of nowhere when considering the ultra-large-scale, high-resolution global medium-term forecasting task. \textsc{Pangu-Weather}\cite{bi2023accurate}, a data-driven model based on 3D Earth-specific transformers, is lauded for its rapid and accurate global forecasts. This model predicts the atmospheric state at a given time based on the current state, described by five upper-air variables on a 0.25° horizontal grid and four surface variables, with 13 vertical levels for the upper-air variables. \textsc{FengWu}~\cite{chen2023fengwu} addresses the medium-range forecasting problem from a multi-modal, multi-task perspective, with its elaborate deep learning architecture with model-specific decoders and cross-modal fusion transformers that learn under the supervision of uncertainty loss to balance the optimization of different predictors in a regionally adaptive manner. FuXi~\cite{chen2023fuxi} cascades cubic embeddings and U-transformers and is trained using 39 years of high-resolution in-analysis data. It delivers forecast performance comparable to that of the ECMWF EM with a temporal resolution of 6 hr and a spatial resolution of 0.25° in a 15-day forecast. The FuXi-Extreme model~\cite{zhong2023fuxiextreme} employs a denoising diffusion probabilistic model (DDPM)~\cite{ho2020denoising} to refine the surface forecast data generated by the FuXi model~\cite{chen2023fuxi} in 5-day forecasts, thereby enhancing extreme rainfall/wind forecasting. As an all-purpose foundation model, \textsc{ClimaX}~\cite{nguyen2023climax} introduces the concept of foundation modeling to the field of weather prediction, with its fully supervised pre-training based on the Transformer, and proposes variable tokenization and variable aggregation strategies for fusing and mining the potential relationships of different weather variations at different heights, which gives it very promising flexibility to adapt to different downstream tasks, including global/regional/seasonal prediction, as well as the tasks of climate mapping, and downscaling. While the aforementioned models are trained in a fully supervised-based pre-training, \textsc{W-MAE}\cite{man2023w} leverages a Masked Auto-Encoder (MAE)-based approach~\cite{he2022masked,feichtenhofer2022masked} for self-supervised training in weather forecasting models, potentially allowing fine-tuning by different data sources to adapt to downstream tasks.

Generative AI are carving a niche in the field of climate and weather forecasting, with several promising approaches recently reported. SEEDS\cite{li2023seeds}, for instance, employs an array of finely-tuned ensemble simulators to generate probabilistic weather forecasts. These forecasts are akin to the ``seeds`` of weather states provided during the inference process, with two different ensemble simulators generating two distinct event predictions. However, the self-regression mechanism underpinning this approach, similar to the RNN architecture used in diffusion model training, is susceptible to instability and feature dissipation over time, particularly in long-range forecasting tasks. Contrastingly, Dyfussion~\cite{cachay2023dyffusion} uses pristine initial conditions, while the PDE-Refiner \cite{lippe2023pde} enhances the diffusion process-based predictions by iteratively observing them to capture low-amplitude information that may not be immediately evident in the data. DITTO~\cite{ovadia2023ditto} adopts a unique approach, generating a continuous interpolation between the initial and final time steps, and using time fireworks instead of incremental noise in the forward process. TemperatureGAN~\cite{balogun2023temperaturegan}, a conditional GAN, considers factors such as the month, location, and time period to generate atmospheric temperature predictions at an hourly resolution above ground level. 

Furthermore, GANs that integrate physical information constraints are being deployed to emulate ocean systems, thereby enhancing climate prediction capabilities \cite{meng2021physics,meng2023physical,lutjens2021physically,yuan2023space,chen2021physics}.For instance, Refs.~\cite{meng2021physics,meng2023physical} describe GAN-based models that learn underlying physical relationships between surface and subsurface temperatures in numerical models. Subsequent calibration of model parameters using observational data leads to enhanced predictions. PGnet \cite{chen2021physics} is a generative neural network model that uses a mask matrix to identify regions of low-quality prediction generated during the initial physical stage. The generative neural network then uses this mask as a prior for the second stage of fine prediction. WGC-LSTM~\cite{wilson2018low} harnesses graph convolutions to capture spatial relationships and amalgamates these with LSTM to concurrently consider both spatial and temporal relationships.

Reflecting upon the intricate interconnections between atmospheric elements, surface variables, and precise terrestrial coordinates within the Earth system, a substantial amount of research has utilized graph-based methodologies for weather and climate prediction tasks. For instance, \textsc{Keisler Graph Neural Network}~\cite{keisler2022forecasting} leverages a graph neural network architecture~\cite{scarselli2008graph} to achieve weather forecasting. It uses an encoder that maps the original 1° latitude/longitude mesh to an icosahedral mesh, performs message passing computations on this mesh, and then decodes back into latitude/longitude space. \textsc{Graphcast}~\cite{lam2022graphcast}, on the other hand, also utilizes a GNN-based framework for weather prediction, albeit with a much higher resolution and flexibility. It stands as the inaugural large-scale foundation model for weather and climate predictions based on graph methodology. \textsc{Graphino}~\cite{cachay2021world}, a globally spatial GNN, is specifically designed for seasonal forecasting tasks, including prediction of the El Nino-Southern Oscillation (ENSO) phenomenon~\cite{wang2022enso}. The model begins by constructing an initial graph with grid cells as nodes and learns the edges based on the connectivity between geographical locations. In addition, GE-STDGN~\cite{ni2022ge} employs a graph structure learning and optimization method underpinned by the evolutionary multi-objective optimization (EMO) algorithm known as graph evolution~\cite{leskovec2007graph}. This augments the model's ability to analyze intricate node correlations for spatio-temporal weather sequence prediction. HiSTGNN~\cite{ma2023histgnn} features an adaptive graph learning module that builds a self-learning hierarchical graph~\cite{ying2018hierarchical}. This graph is comprised of a global graph that represents region-specific information and a local graph that encapsulates meteorological variables within each region. The model effectively identifies hidden spatial dependencies and diverse long-term weather patterns using graph convolution and gated temporal convolution with a dilated initial as its core structure. Lastly, WeKG-MF~\cite{ayadi2022wekg} presents an innovative approach by constructing a knowledge graph from open weather observations published by Météo-France. This model is built upon a semantic schema that encapsulates the knowledge of meteorological observations for an array of downstream scenarios.

\subsection{Precipitation Nowcasting}
The domain of precipitation nowcasting has garnered substantial advancements through the application of DL techniques, including CNNs~\cite{chen2022dynamic,lee2019mcsip,cuomo2021developing,gong2023enhancing, ehsani2022nowcasting,fernandez2021broad}, RNNs~\cite{shi2015convolutional,wu2021motionrnn,huang2022mmstn,dai2022mstcgan,luo2020pfst, ehsani2022nowcasting}, and Transformers~\cite{Chen_2023,dong2022motion,yang2021ptct,bai2022rainformer,jin2023preformer,parknowformer}. These methodologies have demonstrated remarkable proficiency in managing spatio-temporal data, a prevalent format in Earth system observation.

\textsc{ConvLSTM}~\cite{shi2015convolutional} was pioneering in its integration of deep learning for processing precipitation proximity forecasts, effectively amalgamating CNN and LSTM to manage spatio-temporal radar data. Successive models, such as \textsc{PredRNN}~\cite{wang2022predrnn} and \textsc{E3D-LSTM}~\cite{wang2018eidetic}, similarly incorporate spatio-temporal data within LSTM and CNN architectures to extract long-term higher-order correlations. \textsc{PhyDNet}~\cite{guen2020disentangling} introduces partial differential equation (PDE)\cite{evans2022partial} constraints into its theoretical space. \textsc{MetNet}\cite{sonderby2020metnet} and its subsequent iterations, \textsc{MetNet-2} and \textsc{MetNet-3}~\cite{andrychowicz2023deep}, proposed an architecture based on ConvLSTM and advanced CNNs, thereby enabling proficient precipitation forecasting up to 12 hours ahead.

The ascension of Transformers in the visual realm has benefited the spatio-temporal video streaming data-based approach to rainfall prediction. For instance, PTCT~\cite{yang2021ptct} divides original frames into multiple patches to eliminate inductive bias constraints. It also applies 3D temporal convolutions to effectively capture short-term dependencies. The Preformer~\cite{jin2023preformer} model proposes an encoder-translator-decoder architecture where the encoder integrates spatial features from multiple elements, the translator models spatio-temporal dynamics, and the decoder combines temporal and spatial information for future precipitation prediction. Rainformer~\cite{bai2022rainformer} introduces global feature extraction units and gate fusion units (GFUs) to balance the fusion of local and global features, thereby enabling efficient rainfall prediction. \textsc{TempEE}~\cite{Chen_2023} proposes a parallel use of spatio-temporal encoders and decoders based on the Transformer architecture, achieving promising results in the egoless regression strategy for handling non-stationary spatio-temporal sequences. This significantly improves the accuracy of precipitation nowcasting. \textsc{Earthformer} model~\cite{gao2022earthformer}, based on Cuboid Attention, is utilized for Earth system forecasting, including precipitation nowcasting and ENSO~\cite{cai2015enso}. 

Taking into account the instructive role of knowledge from other modes, multimodal spatial-temporal tasks have been introduced~\cite{ma2023mm,jin2023spatiotemporal,huang2022mmstn}. The \textsc{MM-RNN}~\cite{ma2023mm} introduces elemental knowledge to guide precipitation nowcasting, enforcing a constraint that requires the movement of precipitation to follow basic atmospheric laws of motion for accurate forecasting. \textsc{STIN}~\cite{jin2023spatiotemporal} utilizes spatio-temporally specific filters to generate precipitation forecasts from multi-modal meteorological data. Recently, precipitation nowcasting, viewed as an uncertainty assessment problem, has also benefited from the successful application of diffuison modeling.

Recently, precipitation nowcasting, viewed as an uncertainty assessment problem, has also benefited from the successful application of generative modeling. \textsc{DGMR} employs an adversarial training methodology to generate sharp and accurate proximity forecasts, which solves the problem of fuzzy prediction. \textsc{DMSF-GAN}~\cite{chen2022dynamic}, on the other hand, completely eschews autoregressive strategies and is based on adversarial training and pure CNN architectures to address the problem of feature dispersion over time.\textsc{PCT-CycleGAN}~\cite{choi2023pct} generates temporal causality using two generator networks with forward and backward temporal dynamics. Each generator network learns a multitude of one-to-one mappings on precipitation data based on time-dependent radar to approximate a mapping function representing the temporal dynamics in each direction. The MPL-GAN~\cite{liu2020mpl} utilizes a multi-path learning strategy to improve the diversity of generated sequences while providing accurate predictions. In addition, to simultaneously handle uncertainty and enhance domain-specific standards, PreDiff~\cite{gao2023prediff} adopts a two-stage probabilistic spatiotemporal prediction pipeline, incorporating explicit knowledge control mechanisms to enforce predictions conforming to specific domain's physical constraints. This is achieved by estimating the bias of the constraints imposed in each denoising step and correspondingly challenging the overfitting distribution. GED~\cite{asperti2023precipitation}, known as Generative Ensemble Diffusion, utilizes a diffusion model to generate a set of possible weather scenarios which are then amalgamated into a probable prediction via the use of a post-processing network. Ref.~\cite{ravuri2021skilful} utilizes radar-based deep learning models for skillful short-term precipitation forecasting, achieving display-consistent predictions over a 1536x1280 km region. 

The introduction of physical constraints and graph relations can improve the efficiency and accuracy of the model. Ref.~\cite{hess2022physically} introduces a generative adversarial network with physical information constraints to improve both local distribution and spatial structure for daily precipitation field improvement. CNGAT~\cite{peng2021cngat} fuses spatial and temporal information for improved Radar quantitvative precipitation estimate (RQPE)~\cite{zhang2016multi}. The precipitation estimation area was partitioned into subares that were treated as nodes to from an input graph. All nodes were then categorized according to the temporal mean radar reflectivity for precipitation estimation with an attention mechanism.

\subsection{Downscaling}
Achieving precise, fine-grained weather predictions necessitates high spatial resolution data. However, most global weather forecasting models are restricted by the availability and scale of data, resulting in an over-reliance on data with approximately a 5.625° spatial resolution, equivalent to a grid point spacing of about 625 kilometers. Despite these limitations, the data volume is significant. For instance, the data scale of the ERA5 system at a 0.25° spatial resolution is several tens of times larger—around 15 terabytes—compared to the 5.625° spatial resolution data. High spatial resolution data offer a more granular representation of complex atmospheric processes and the interplay between different weather systems. One strategy to tackle this issue is the enhancement of weather data's spatial resolution, a process referred to as super-resolution (SR)~\cite{sharma2022resdeepd,ballard2022contrastive}. SR can bolster the resolution of gridded data, surpassing conventional interpolation methods in effectiveness.

A popular DL-based SR model, U-Net, leverages a synergistic encoder-decoder structure to produce high-resolution outputs from low-resolution inputs~\cite{hu2019runet,min2023d,yu2023weather}. Within the realm of semi-supervised learning, generative adversarial networks (GANs) have demonstrated potential in enhancing the representation of more intricate structures and details~\cite{harris2022generative,annau2023algorithmic,mardani2023generative,leinonen2020stochastic,wang2018esrgan,watson2020investigating,wang2021fast,stengel2020adversarial}. The typical procedure involves training the generator to learn the potential mapping between low- and high-resolution grid data or images. For example, Stengel et al. presented an adversarial DL approach that super-resolves the predictions of wind speed and solar irradiance in global climate models to a sufficient scale for renewable energy resource assessment, thereby improving the resolution of wind and solar energy data nearly fiftyfold~\cite{stengel2020adversarial}.

Recent research has deployed diverse strategies such as normalizing flows and neural operators. Self-supervised learning-based methods have also been investigated for downscaling low-resolution grid weather data. For instance, the pre-trained foundation model, \textsc{ClimaX}~\cite{nguyen2023climax}, allows fine-tuning for resolution downscaling. González et al. introduced a downscaling strategy based on multi-variable physical hard constraints, ensuring the physical relationships between variable sets~\cite{gonzálezabad2023multivariable}. 

Physics-constrained DL-based methods have also been proposed to improve the model's performance via external adjustment~\cite{harder2023physics,juan2023data,feng2023physics,fuchs2023torchclim,hammoud2022cdanet,harder2022generating, esmaeilzadeh2020meshfreeflownet}. For example, MeshfreeFlowNet~\cite{esmaeilzadeh2020meshfreeflownet} employs a physics-informed model which incorporates Partial Differential Equations (PDEs) as regularization terms into the loss function, achieving spatio-temporal downscaling. Harder et al.~\cite{harder2022generating} were the first to apply hard-constraining to achieve fine-grained downscaling outputs in climate change datasets. Furthermore, strategies such as contrastive learning~\cite{ballard2022contrastive} and Betrays DL models~\cite{gerges2022novel} were adpoted. 

In response to the lack of interpretability of DL-based downscaling methods, Gong et al. explored the interpretability of fundamental CNNs in climate model downscaling strategies, thus paving the way for trustworthy artificial intelligence in downscaling models~\cite{gong2023enhancing}. Bano et al. analyzed the downscaling issue from a multimodel perspective, developing a CNN-based downscaling prediction ensemble (DeepESD) for temperature and precipitation in the European EUR-44i (0.5°) domain based on eight global circulation models~\cite{bano2022downscaling}. This represents the first application of CNNs in generating a downscaled multimodel ensemble based on perfect prognosis methods, allowing for the quantification of model uncertainty in climate change signals.

The introduction of uncertainty modeling also allows downscaling gains in DL-based models, significantly improving efficiency as well as reconstruction resolution. ResDiff~\cite{mardani2023generative} employs a two-step diffusion model-based approach. In the first step, U-Net regression predicts the mean values, while in the second step, the diffusion model predicts the residuals, thereby achieving kilometer-scale atmospheric downscaling. However, it should be noted that the use of diffusion models in the field of weather and climate is still in the exploratory stage. Ref.~\cite{hatanaka2023diffusion} also employs similar operations, utilizing diffusion models for cloud cover and super-resolution diffusion models for high-resolution solar energy forecasting. 

\subsection{Bias Correction}
Bias correction in weather predictions has traditionally relied on statistical methods~\cite{yoshikane2022bias}. Over time, these techniques have evolved, embracing machine learning strategies such as Deep Belief Networks and Support Vector Machines. The advent and proliferation of data availability have further catalyzed the shift towards deep learning methodologies, including Long Short-Term Memory (LSTM)~\cite{li2022numerical,yang2022correcting,blanchard2022multiscale} and Convolutional Neural Networks (CNN)~\cite{han2021deep,han2022short,wang2022deep}. These methodologies have been instrumental in mitigating common weather-related biases.

A notable approach is the DL-Corrector-Remapper technique~\cite{ge2022dl}, which stands apart in its ability to correct, remap, and fine-tune gridded uniform forecasts from the FourCastNet system. This process enables a direct comparison with non-uniform, sparse observational ground truth data via the AFNO method. The Super Resolution Deep Residual Network (SRDRN)~\cite{wang2022deep} has been employed for climate downscaling and bias correction. This network utilizes stacked general circulation models and extracts spatial features, effectively diminishing biases and correcting spatial dependencies relative to observational data.

In an intriguing application, the Unsupervised Image-to-Image Translation (UNIT) network~\cite{fulton2023bias} capitalizes on unpaired image translation for bias correction. This method offers a novel perspective on bias mitigation. Hess et al.~\cite{hess2022physically} have proposed a post-processing technique that employs a physics-constrained Generative Adversarial Network (cGAN) to concurrently correct biases in local frequency distribution and spatial patterns of state-of-the-art CMIP6-level Earth System Models.

Recently, the WeatherGNN model~\cite{wu2023weathergnn} has been developed, which leverages a Graph Neural Network within a comprehensive framework. This model learns the intricate relationships between weather and geography, capturing meteorological interactions and spatial dependencies between grids. This approach provides a robust and sophisticated tool for bias correction. These advancements illustrate the potential of deep learning methodologies in refining weather prediction systems.

\subsection{Data Assimilation}
Data assimilation (DA) is a key component of high-level NWP systems. These systems not only forecast future states, but also integrate observational data to establish the initial state, guiding the model's trajectory to future states. This complex process is computationally demanding, making it an active area of research. Existing approaches often rely on simplifying assumptions, such as linearity, which adds to the challenges in the field. However, the integration of deep learning into DA is gaining recognition, with encouraging research outcomes. For instance, \textsc{Oceanfourcast}\cite{bire2023oceanfourcast} employs neural operators alongside a Transformer-based architecture, inspired by \textsc{FourCastNet}\cite{pathak2022fourcastnet}, to support adjoint-based data assimilation in ocean modeling. Furthermore, Bocquet et al.\cite{Bocquet_2020} innovatively combine DA, machine learning, and expectation maximization to perform Bayesian inference of chaotic dynamics, enabling the assimilative reconstruction of observational data for geophysical flows. For an in-depth review of DA, we refer readers to Geer's work\cite{geer2021learning}.

\subsection{Climate Text Analysis}
The rapid development of LLMs has provided new insights for climate text analysis. Hershcovich et al. introduced a climate performance model card, designed with the intent of practical application requiring minimal information on the experimental setup and associated computer hardware~\cite{hershcovich2022towards}. A language model, known as \textsc{ClimateBert}~\cite{webersinke2022climatebert}, was developed with a foundation in \textsc{DistillRoBERTa}, specifically designed for analyzing climate-orientated text. This versatile model can be employed in a variety of tasks, such as detecting climate-related content, discerning sentiment in climate-related paragraphs, identifying commitment and action-related content, distinguishing specific from non-specific climate-related text, and assigning climate-related content to one of four categories as per the recommendations of the Task Force on Climate-related Financial Disclosures (TCFD Further refinement of \textsc{ClimateBert} is seen in the work of Garrido-Merchán et al.~\cite{garridomerchán2023finetuning}, who utilized ClimaText~\cite{varini2021climatext} to fine-tune the model for the specific task of analyzing disclosures relating to financial risks connected with climate change. An extension, \textsc{ClimateBERT-NetZero}~\cite{schimanski2023climatebertnetzero}, was designed to classify whether a given text contains a net zero or reduction target. Krishnan et al. employed \textsc{ClimateBERT} in their \textsc{ClimateNLP} project, analyzing public sentiment towards climate change using data gathered from Twitter and Facebook~\cite{krishnan2023climatenlp}. Auzepy et al. proposed the use of pretrained LLMs' zero-shot capabilities to evaluate TCFD reporting~\cite{auzepy2023evaluating} . However, this approach is not without its challenges. Pre-trained LLMs often lack up-to-date information and tend to use imprecise language, a significant disadvantage in the field of climate change where accuracy is paramount. To mitigate this, Kraus et al. incorporated emission data from ClimateWatch and utilized a general Google search to enhance the language model~\cite{kraus2023enhancing}. Vaghefi et al. integrated information from the Intergovernmental Panel on Climate Change's Sixth Assessment Report (IPCC AR6) into GPT-4~\cite{openai2023gpt4}, laying the groundwork for the implementation of conversational AI in the realm of climate science~\cite{vaghefi2023chatclimate} . In the intersection of climate and health, \textsc{CliMedBERT}~\cite{fard2022climedbert} was developed for diverse applications, including understanding climate and health-related concepts, fact-checking, relationship extraction, and generating evidence on the impact of health on policy text generation. Additionally, Bi et al. have proposed OceanGPT~\cite{bi2023oceangpt}, based on LLM (e.g., Llama~\cite{touvron2023llama} and GPT3.5~\cite{brown2020language}), to handle specific tasks related to the ocean, such as ocean text analysis and intelligent underwater agent instructions.

\subsection{Weather Patterns Understanding}
Weather pattern understanding, as opposed to forecasting, tends to lean towards a qualitative analysis of climate change. By integrating predictions derived from reanalysis datasets, we can more effectively quantify the potential impact of future weather events. Traditional numerical methods, though costly, rely on manually crafted features such as fronts, tropical cyclones~\cite{knutson2020tropical}, extratropical cyclones, and atmospheric rivers, using heuristic detection algorithms based on empirical knowledge. However, weather patterns with more distinct features, like tornadoes and typhoons, may be more amenable to pattern detection and prediction due to their characteristic features. For instance, a typhoon's eye and surrounding rainbands present distinct patterns. This pattern detection and prediction could potentially prove more advantageous than predicting the general atmospheric state in standard training. One approach might be to employ spatio-temporal video stream data, such as radar reflectivity data~\cite{Chen_2023} and weather satellite cloud imagery~\cite{bai2023lsdssimr}. This transition from spatio-temporal weather video stream data to predictions offers a more dynamic and visually intuitive method for weather pattern understanding.

Weather pattern understanding based on DL techniques often requires large-scale, well-annotated samples. In one study, Kashinath et al.\cite{kashinath2021climatenet} created a dataset suitable for tropical cyclone (TC) detection in the 25km CAM5.1 model. They achieved fine-grained and rapid segmentation of TCs and atmospheric rivers (ARs) using DL-based segmentation algorithms. Racah et al.\cite{racah2017extremeweather} extended this dataset to detect and precisely locate TCs, extra-tropical cyclones (ERCs), ARs, and tropical low-pressure systems using a 3D-CNN. Furthermore, Sobash et al.~\cite{sobash2023diagnosing} combined CNNs and logistic regression (LR) to detect tornadoes in six-hourly dynamical forecasts and turbulence conditions in regional or high-resolution weather forecasts. In addition to detecting different weather patterns from large-scale reanalysis datasets, advanced AI models are frequently used to study the evolutionary processes of meteorological phenomena. These include the genesis and dissipation of typhoons in a regional context, as well as the movement trajectories of TCs. Next, we will proceed to conduct a literature review and discussion on the field of weather pattern understanding, focusing on climate phenomena and extreme weather events.

\subsubsection{Climate Phenomena Understanding and Prediction}
We mainly focus on discussing three primary climate phenomena/representation in the global scale, including El Niño-Southern Oscillation, Climate Tipping Points, and Madden-Julian Oscillation.

\begin{itemize}
    \item \textbf{El Niño-Southern Oscillation.} The El Niño phenomenon, arising from intense ocean-atmosphere interactions, is marked by heightened sea surface temperatures (SST), a levelled equatorial Pacific thermocline, and a diminished tropical Pacific Walker circulation~\cite{rasmusson1982variations}. Together with its inverse phase, La Niña, it constitutes the El Niño-Southern Oscillation (ENSO) cycle. This cycle, with a duration of 2 to 7 years, is the principal driver of global climate interannual variability, frequently correlating with significant global climatic and socio-economic repercussions~\cite{latif1998review}. Consequently, accurate ENSO forecasting is of paramount scientific and practical significance. Several methods have been proposed to enhance ENSO forecasting. Ref.\cite{ye2021transformer} incorporates a Transformer-based architecture, considering long-term correlations among meteorological variables. A spatial-temporal Transformer for multi-year ENSO prediction is suggested by Ref.\cite{song2023spatial}. ENSO-GTC~\cite{mu2022enso} applies the Global Teleconnections Coupler (GTC) for potential teleconnections between global SST. Ref.\cite{wang2023interpretable,liu2023explainable} develop an interpretable deep learning model for ENSO forecasting. Ref.\cite{ham2021unified} introduces a holistic deep learning model for ENSO that integrates seasonality in climate data to enhance forecast fluctuation. Comprehensive reviews and surveys on deep learning-based ENSO forecasting can be found in Refs.~\cite{fang2022survey,wang2022enso}.

    \item  \textbf{Climate Tipping Points.} Climate tipping points denote crucial thresholds within the climate system where the system undergoes significant and irreversible alterations in response to certain changes or external forcings~\cite{liuschiaffini2023tipping,andrychowicz2023deep,gnanadesikan2023using}. These transitions can instigate major climate system shifts, including modifications in oceanic circulation patterns, accelerated glacier melting, and climate zone migration. The transgression of these tipping points can destabilize the long-term equilibrium of the climate system, inciting more severe climate transformations. \textsc{TIP-GAN}\cite{sleeman2023generative} is a Generative Adversarial Network (GAN)-based model designed to identify potential climate tipping points in Earth system models, with a particular emphasis on precipitating the collapse of the Atlantic Meridional Overturning Circulation (AMOC). Additionally, a neural-symbolic question answering program translator, NS-QAPT, is presented as a neural-symbolic approach to enhance the interpretability and explainability of deep learning climate simulations applied to climate tipping point detection\cite{andrychowicz2023deep}. Further relevant works can be explored in Refs.~\cite{rietkerk2021evasion,bury2021deep}.

    \item \textbf{Madden-Julian Oscillation.} The Madden-Julian Oscillation (MJO)~\cite{zhang2005madden,zhang2013madden} is a substantial atmospheric circulation phenomenon predominantly observed near the equator. It is characterized by regular oscillations in convection activity and precipitation in equatorial regions, with a typical duration spanning 20 to 90 days. The MJO exerts substantial influence on global weather and climate systems, impacting precipitation patterns, wind fields, and the origination and evolution of tropical cyclones. Consequently, comprehension and prediction of the MJO are vital for accurate precipitation forecasting and disaster prevention, thereby effectively managing and mitigating potential risks. DK-STN~\cite{li2023dk}, leveraging spatio-temporal knowledge embedding, has notably enhanced the prediction accuracy of the ANN method, while preserving high levels of efficiency and stability. For further related works, refer to Ref.~\cite{materia2023artificial}.
\end{itemize}

\subsubsection{Extreme Weather Prediction and Understanding}
This discussion primarily centers around the application of DL models for the prediction and understanding of four pivotal extreme weather events: Extreme Temperatures, Drought, Cyclones, and Extreme Precipitation.

\begin{itemize}
    \item \textbf{Extreme temperatures.} \textbf{Extreme Temperatures.} These often present as intense, prolonged, and frequent heatwaves~\cite{jacques2022deep}, imposing substantial challenges to human activities and the ecological environment. Extreme temperature events are typically defined as a series of days with temperature variables surpassing a specific threshold or evaluated using accumulation indices composed of amplitude, duration, and frequency. Data-driven climate models rooted in machine learning/deep learning have demonstrated effectiveness in extreme temperature prediction tasks. Techniques such as random forest and XGBoost have offered promising results. Furthermore, convolutional neural networks, recurrent neural networks, and Transformers have seen extensive use in extreme temperature prediction due to their capacity to capture spatiotemporal representations.
    
    \item \textbf{Drought.} Droughts occur at various spatiotemporal scales and involve multiple triggering mechanisms, which complicates a clear and comprehensive definition~\cite{minixhofer2021droughted}. They represent an extremely complex natural disaster. Recent research has gravitated towards using AI algorithms~\cite{grabar2023long} based on geospatial weather data for long-term drought prediction, such as in \cite{ahmed2021hybrid, danandeh2023novel,xu2022application}. For example, Ref.\cite{ahmed2021hybrid} proposed a one-dimensional CNN combined with a GRU for evapotranspiration prediction, enabling the model to better capture dependencies in time series data. Meanwhile, Ref.\cite{danandeh2023novel} combined CNN and LSTM for drought prediction one month in advance. A more comprehensive review of AI applications in drought prediction can be found in Refs.\cite{ahmed2021hybrid,prodhan2022review}. However, most existing studies are geographically focused, causing the model performances to heavily depend on specific research conditions such as the study area, drought index, or considered input variables. This dependency makes it difficult to generalize major findings from one study to another.
    
    \item \textbf{Cyclones \& Extreme Precipitation.} In tropical and mid-latitude regions, weather-scale cyclones represent some of the most extreme events causing significant economic damage due to heavy rainfall, strong winds, and storm surges~\cite{mendelsohn2012impact}. Evidence suggests that climate change may amplify the severity of these extreme events, even if not their frequency~\cite{knutson2020tropical}. However, predicting their variability on sub-seasonal to decadal timescales remains a challenge~\cite{befort2022seasonal}. Heavy precipitation events are not always linked with large-scale weather systems such as cyclones or fronts; many impactful events are tied to brief, small-scale severe convective events. These extremes pose a greater challenge for operational climate prediction systems as their spatial resolution is too coarse to capture the explicit representation of convection. In most regions where extreme precipitation is analyzed, the skill of numerical climate prediction systems for extreme precipitation decreases significantly after a few days. AI techniques have been applied to improve the prediction of cyclones and heavy precipitation events from various perspectives. The objective is to enhance the skill of numerical prediction systems (e.g., seasonal forecasting) in representing extreme weather events by identifying the relationship between large-scale driving factors and the occurrence of extreme events. This approach has been applied to large-scale extreme events, such as tropical or extratropical cyclones, or directly to precipitation fields~\cite{scheuerer2020using,specq2020improving}. De Burgh-Day \& Leibnberg~\cite{de2023machine} proposed a systematic model ablation study as a potential approach to address the interpretability issue of DL models while maintaining their good skill. Additionally, some DL-based strategies aim to handle cyclones and extreme precipitation forecasting via meteorological image extrapolation (refer to Section. \textbf{Precipitation Nowcasting}), and others focus on improving model outputs by achieving high-resolution observations to enhance the representation of precipitation or wind patterns associated with cyclones rather than directly performing the prediction task~\cite{yang2022machine,vosper2023deep} (refer to Sec.~\ref{sec:applications}).
\end{itemize}

\section{Resources}
\label{sec:dataset}
In this section, we catalog the prevalent datasets and tools pertinent to weather and climate change analysis, aspiring to streamline their accessibility for practitioners.

\subsection{Dataset}
This segment classifies datasets employed in data-driven weather and climate studies. These datasets facilitate weather time-series analysis, weather spatio-temporal series analysis, weather spatio-temporal video stream analysis, and climate text analysis. We bifurcate them into two categories: weather and climate series data and climate text data. It's noteworthy that the datasets are unordered.
\begin{table*}[tbh]
  \centering
  \caption{Summary of weather and climate-related dataset resources in different applications. (\textbf{FO}: Forecasting; \textbf{PR}: Projection; \textbf{DO}: Downscaling; \textbf{BC}: Bias Correlation; \textbf{DA}: Data Assimilation; \textbf{WPU}: Weather Pattern Understanding; \textbf{PN}: Precipitation Nowcasting; \textbf{CTA}: Climate Text Analysis. \textbf{All datasets have accessible hyperlinks on their names.}}
  \resizebox{1\textwidth}{!}{
    \begin{tabular}{crcccrrrrrrrr}
    \toprule
    \multicolumn{2}{c}{\multirow{2}[2]{*}{\textbf{Data Types}}} & \multirow{2}[2]{*}{\textbf{Dataset}} & \multirow{2}[2]{*}{\textbf{Statistics}} & \multirow{2}[2]{*}{\textbf{Timeframe}} & \multicolumn{8}{c}{\textbf{Applications}} \\
    \multicolumn{2}{c}{} &    &    &    & \multicolumn{1}{c}{FO} & \multicolumn{1}{c}{PR} & \multicolumn{1}{c}{DO} & \multicolumn{1}{c}{BC} & \multicolumn{1}{c}{DA} & \multicolumn{1}{c}{WPU} & \multicolumn{1}{c}{PN} & \multicolumn{1}{c}{CTA} \\
    \midrule
    \multirow{29}[6]{*}{Time Series} & \multicolumn{1}{c}{\multirow{8}[2]{*}{Reanalysis/Simulation}} & \href{https://github.com/pangeo-data/WeatherBench}{CMIP6}~\cite{rasp2020weatherbench,rasp2023weatherbench,nguyen2023climatelearn} & Reanalysis Grid Data & 1850-2100 &  \Checkmark  &  \Checkmark  &  \Checkmark  &  \Checkmark  &  \Checkmark  &  \XSolidBrush   &  \XSolidBrush  &  \XSolidBrush\\
       &    & \href{https://github.com/pangeo-data/WeatherBench}{ERA5}~\cite{rasp2020weatherbench,rasp2023weatherbench,nguyen2023climatelearn} & Reanalysis Grid Data & 1779 to the present &  \Checkmark  &  \Checkmark  &  \Checkmark  &  \Checkmark  &  \Checkmark  &  \XSolidBrush   &  \XSolidBrush  &  \XSolidBrush\\
       &    & \href{https://tianchi.aliyun.com/dataset/98942}{HCOSD} & Reanalysis Grid Data & 1850-2100 &  \Checkmark  &  \Checkmark  &  \XSolidBrush  &  \XSolidBrush  &  \XSolidBrush  &  \Checkmark  &  \XSolidBrush  & \XSolidBrush\\
       &    & \href{https://github.com/aditya-grover/climate-learn}{Extreme-ERA5}~\cite{nguyen2023climatelearn} & Reanalysis Grid Data & 1979-2018 &  \Checkmark  &  \Checkmark  &  \Checkmark  &  \Checkmark  &  \Checkmark  &  \Checkmark  &  \XSolidBrush  &  \XSolidBrush\\
       &    & \href{https://extremeweatherdataset.github.io/}{ExtremeWeather}~\cite{racah2017extremeweather} & Reanalysis Grid Data & 1979-2005 &  \Checkmark  &  \Checkmark  & \XSolidBrush   &  \XSolidBrush  &  \XSolidBrush  &  \XSolidBrush  &  \Checkmark  &  \XSolidBrush\\
       &    & \href{https://portal.nersc.gov/project/ClimateNet/}{ClimateNet}~\cite{kashinath2021climatenet} & Reanalysis Grid Data & 1996-2010 &  \Checkmark  &  \Checkmark  &  \XSolidBrush  &   \XSolidBrush &   \XSolidBrush &  \Checkmark  &  \XSolidBrush  &  \XSolidBrush\\
       &    & \href{https://github.com/spcl/ens10}{ENS-10}~\cite{ashkboos2022ens} & Reanalysis Grid Data & 1998-2017 &  \Checkmark  &  \XSolidBrush  &  \Checkmark  &  \Checkmark  &  \Checkmark  & \XSolidBrush   &  \XSolidBrush  & \XSolidBrush \\
       &    & \href{https://github.com/RolnickLab/climart}{ClimART}~\cite{cachay2021climart} & Reanalysis Grad Data & 1979-2014 &  \Checkmark  &  \Checkmark  &  \XSolidBrush & \Checkmark  &  \Checkmark  &  \Checkmark  &  \XSolidBrush &  \XSolidBrush\\
\cline{2-13}       & \multicolumn{1}{c}{\multirow{11}[2]{*}{Observation}} & \href{https://zenodo.org/records/3114194}{China-Precipitation},\href{https://zenodo.org/records/3185722}{Temperature}~\cite{peng20191} & Observation Data & 1901-2017 &  \Checkmark  &  \XSolidBrush  &  \XSolidBrush  & \XSolidBrush   &  \XSolidBrush  &  \XSolidBrush  &  \XSolidBrush  &  \XSolidBrush\\
       &   & \href{https://github.com/kitamoto-lab/digital-typhoon/}{Digital Typhoon}~\cite{kitamoto2023digital} & Observation Data & 1978-2022 &  \Checkmark  &  \XSolidBrush  & \XSolidBrush   &  \XSolidBrush  &  \XSolidBrush  &  \Checkmark  &  \Checkmark  &  \XSolidBrush \\
       &    & \href{https://www.kaggle.com/datasets/cdminix/us-drought-meteorological-data}{DroughtED}~\cite{minixhofer2021droughted} & Observation Data & June 2017 - December 2017 &  \Checkmark  &  \XSolidBrush  & \XSolidBrush   &  \XSolidBrush  &  \XSolidBrush  &  \Checkmark  &  \XSolidBrush  &  \XSolidBrush \\
       &    & \href{https://github.com/uihilab/IowaRain}{IowaRain}~\cite{sit2021iowarain} & Observation Data & 2016-2019 &  \Checkmark  &  \Checkmark  &  \XSolidBrush  &  \XSolidBrush  &  \XSolidBrush  &  \Checkmark  &  \XSolidBrush  &  \XSolidBrush\\
       &    & \href{https://tianchi.aliyun.com/competition/entrance/231662/information}{SRAD2018} & Observation Data & 2018 &  \XSolidBrush  &  \XSolidBrush  &   \XSolidBrush &  \XSolidBrush  &  \XSolidBrush  &  \Checkmark  &  \Checkmark  &  \XSolidBrush\\
       &    & \href{https://drive.google.com/file/d/1R6hS5VAgjJQ_wu8i5qoLjIxY0BG7RD1L/view}{KnowAir}~\cite{wang2020pm2} & Observation Data & 2015-2018 &   \Checkmark  &    \XSolidBrush &   \XSolidBrush  &  \XSolidBrush   &   \XSolidBrush  &   \XSolidBrush  &   \XSolidBrush  &  \XSolidBrush\\
       &    & \href{https://data.nasa.gov/}{NASA}~\cite{chen2023prompt} & Observation Data & 2012-2022 & \Checkmark  &  \XSolidBrush  &  \XSolidBrush  & \XSolidBrush   & \XSolidBrush   &  \XSolidBrush  &  \XSolidBrush  & \XSolidBrush \\
       &    & \href{https://github.com/aditya-grover/climate-learn}{PRISM}~\cite{nguyen2023climatelearn} & Observation Data & 1895 to the present &  \Checkmark  &  \Checkmark  &  \Checkmark  &  \Checkmark  &  \Checkmark  &  \Checkmark  &  \XSolidBrush  &  \XSolidBrush\\
       &    & \href{https://neuralchen.github.io/RainNet/}{RainNet}~\cite{chen2020rainnet} & Observation Data & None &  \XSolidBrush   &   \XSolidBrush  &  \Checkmark  &   \XSolidBrush  &   \XSolidBrush  &  \Checkmark  &  \Checkmark  &  \XSolidBrush\\
       &    & \href{https://data.caltech.edu/records/grs4f-c9r41}{Continental United States Wind Speeds}~\cite{kurinchi-vendhan_2021}  & Observation Data & 2007-2013 & \XSolidBrush   & \XSolidBrush   &  \Checkmark  &  \XSolidBrush  &  \XSolidBrush  & \XSolidBrush   & \XSolidBrush   & \XSolidBrush \\
       &    & \href{https://data.caltech.edu/records/ptdmn-3xg52}{Continental United States Solar Irradiance}~\cite{kurinchi-vendhan_2021} & Observation Data & 2007-2013 &  \XSolidBrush  & \XSolidBrush   &  \Checkmark  &  \XSolidBrush  &  \XSolidBrush  & \XSolidBrush   & \XSolidBrush   &\XSolidBrush\\
\cline{2-13}       & \multicolumn{1}{c}{\multirow{11}[2]{*}{Multimodal}} & \href{www.earthnet.tech}{EarthNet2021}~\cite{requena2021earthnet2021} & Multimodal Observation Data & 2018 &  \Checkmark  &  \Checkmark  &  \XSolidBrush  &  \XSolidBrush  &  \XSolidBrush  &  \Checkmark  &  \XSolidBrush  & \XSolidBrush \\
       &    & \href{https://github.com/osilab-kaist/KoMet-Benchmark-Dataset}{KoMet}~\cite{kim2022benchmark} & Multimodal Observation Data & 2011-2018 &  \Checkmark  &  \XSolidBrush  &  \XSolidBrush  &  \XSolidBrush  &  \XSolidBrush  &  \XSolidBrush  &  \XSolidBrush  &  \XSolidBrush \\
       &    & \href{https://www.osti.gov/etdeweb/biblio/21144578}{Germany}~\cite{paulat2008gridded} & Multimodal Observation Data & 2011-2018  &  \Checkmark  &  \XSolidBrush  & \XSolidBrush   &  \XSolidBrush  &  \XSolidBrush  &  \Checkmark  &  \Checkmark  &  \XSolidBrush \\
       &    & \href{https://arxiv.org/pdf/2310.02676.pdf}{China}~\cite{tang2023postrainbench}  & Multimodal Observation Data & 2020-2021  &  \Checkmark  &  \XSolidBrush  & \XSolidBrush  &  \XSolidBrush  &  \XSolidBrush  &  \Checkmark  &  \Checkmark  &  \XSolidBrush\\
       &    & \href{https://www.kaggle.com/datasets/katerpillar/meteonet}{MeteoNet}~\cite{larvor2020meteonet} & Multimodal Observation Data & 2016 to 2018 &  \Checkmark  &  \XSolidBrush  & \XSolidBrush   &  \XSolidBrush  &  \XSolidBrush  &  \Checkmark  &  \Checkmark  &  \XSolidBrush \\
       &    & \href{https://ieeexplore.ieee.org/document/9555094/figures}{RAIN-F}~\cite{choi2021rainfp} & Multimodal Observation Data & 2017-2019 &  \Checkmark  &  \XSolidBrush  &  \XSolidBrush  &  \XSolidBrush  &   \XSolidBrush &   \Checkmark & \Checkmark   &  \XSolidBrush \\
       &    & \href{https://dataon.kisti.re.kr/}{RAIN-F+}~\cite{choi2021rainfp} & Multimodal Observation Data & 2017-2019 &  \Checkmark  &  \XSolidBrush  &  \XSolidBrush  &  \XSolidBrush  &   \XSolidBrush &   \Checkmark & \Checkmark   &  \XSolidBrush\\
       &    & \href{https://sevir.mit.edu}{SEVIR}~\cite{veillette2020sevir} & Multimodal Observation Data & None &  \Checkmark  &  \Checkmark  &  \XSolidBrush  &  \XSolidBrush  &  \Checkmark  &  \Checkmark  &  \Checkmark  &  \Checkmark \\
       &    & \href{https://github.com/frontierdevelopmentlab/pyrain}{RainBench}~\cite{de2021rainbench} & Multimodal Observation \& Reanalysis Data & 2000-2017 &  \Checkmark  &  \Checkmark  &  \Checkmark  &  \XSolidBrush  &  \XSolidBrush  &  \Checkmark  &  \Checkmark  & \XSolidBrush \\
       &    & \href{https://github.com/bycnfz/weather2k}{Weather2K}~\cite{zhu2023weather2k} & Multimodal Observation Data & 2017-2021 &  \Checkmark  &  \XSolidBrush  &  \XSolidBrush  &  \XSolidBrush  &  \XSolidBrush  & \XSolidBrush   &  \XSolidBrush  &  \XSolidBrush\\
       &    & \href{https://ieee-dataport.org/documents/lsdssimr-large-scale-dust-storm-database-based-satellite-images-and-meteorologicall}{LSDSSIMR}~\cite{bai2023lsdssimr} & Multimodal Observation \& Reanalysis Data & 2020-2022 &  \Checkmark  &  \Checkmark  &  \XSolidBrush  &  \XSolidBrush  &  \XSolidBrush  &  \Checkmark  &  \XSolidBrush  &  \XSolidBrush\\
    \midrule
    \multicolumn{2}{c}{\multirow{7}[2]{*}{Text}} & \href{https://huggingface.co/datasets/climate_fever}{CLIMATE-FEVER}~\cite{diggelmann2021climatefever} & Climate-related Text & None & \XSolidBrush   &  \XSolidBrush  &  \XSolidBrush  &  \XSolidBrush  &  \XSolidBrush  &  \XSolidBrush  &  \XSolidBrush  & \Checkmark \\
    \multicolumn{2}{c}{} & ClimateBERT-NetZero~\cite{schimanski2023climatebertnetzero} & Climate-related Text & None & \XSolidBrush   &  \XSolidBrush  &  \XSolidBrush  &  \XSolidBrush  &  \XSolidBrush  &  \XSolidBrush  &  \XSolidBrush  & \Checkmark \\
    \multicolumn{2}{c}{} & \href{https://huggingface.co/datasets/mwong/climatetext-evidence-related-evaluation}{ClimaText}~\cite{varini2021climatext} & Climate-related Text & None & \XSolidBrush   &  \XSolidBrush  &  \XSolidBrush  &  \XSolidBrush  &  \XSolidBrush  &  \XSolidBrush  &  \XSolidBrush  & \Checkmark \\
    \multicolumn{2}{c}{} & \href{https://github.com/climabench/climabench}{CLIMA-INS}~\cite{laud2023climabench} & Climate-related Text & 2012-2021 & \XSolidBrush   &  \XSolidBrush  &  \XSolidBrush  &  \XSolidBrush  &  \XSolidBrush  &  \XSolidBrush  &  \XSolidBrush  & \Checkmark \\
    \multicolumn{2}{c}{} & \href{https://github.com/climabench/climabench}{CLIMA-CDP}~\cite{laud2023climabench} & Climate-related Text & 2012-2021 & \XSolidBrush   &  \XSolidBrush  &  \XSolidBrush  &  \XSolidBrush  &  \XSolidBrush  &  \XSolidBrush  &  \XSolidBrush  & \Checkmark \\
    \multicolumn{2}{c}{} & \href{https://huggingface.co/datasets/iceberg-nlp/climabench}{CLIMATESTANCE \& CLIMATEENG}~\cite{vaid2022towards} & Climate-related Text & None & \XSolidBrush   &  \XSolidBrush  &  \XSolidBrush  &  \XSolidBrush  &  \XSolidBrush  &  \XSolidBrush  &  \XSolidBrush  & \Checkmark \\
    \multicolumn{2}{c}{} & \href{https://huggingface.co/datasets/iceberg-nlp/climabench}{SCIDCC}~\cite{mishra2021neuralnere} & Climate-related Text & None & \XSolidBrush   &  \XSolidBrush  &  \XSolidBrush  &  \XSolidBrush  &  \XSolidBrush  &  \XSolidBrush  &  \XSolidBrush  & \Checkmark \\
    \bottomrule
    \end{tabular}}
  \label{tab:dataset}%
\end{table*}%

\subsubsection{Weather and Climate Series Data}
This subsection concentrates on datasets related to weather and climate sequences, encompassing time series, spatio-temporal sequences, spatio-temporal video streams, and multimodal sequence data.

\textbf{CMIP6}~\cite{rasp2020weatherbench,rasp2023weatherbench, nguyen2023climatelearn} is a compendium of simulated data from Phase 6 of the Coupled Model Comparison Project (CMCP). It encompasses a wide array of different climate variables within the Earth system, such as precipitation, temperature, evapotranspiration, and others. The data, derived from over 150 climate models, spans more than 150 years (1850-2015). It can be utilized to predict the ENSO phenomenon and common climate variables.

\textbf{ERA5}\cite{rasp2020weatherbench,rasp2023weatherbench,nguyen2023climatelearn} is widely used for training and benchmarking data-driven weather and climate forecasting, down-scaling, and projection models. Managed by the European Center for Medium-Range Weather Forecasting (ECMWF)~\cite{trenberth1988evaluation}, it is regularly updated. ERA5 contains hourly data on a 0.25° grid from 1979 till present, at 37 different pressure levels, as well as various surface climate variables, resulting in nearly 400,000 data points at a resolution of 721 × 1440.

\textbf{HCOSD\footnote{\url{https://tianchi.aliyun.com/dataset/98942}}} is provided by the Institute for Climate and Applied Frontier Research (ICAR), is a refined subset of the CMIP dataset. Standing for Historical Climate Observation and Stimulation Dataset, it includes historical simulated data from the CMIP5/6 model and assimilated data from nearly a century of historical observations, reconstructed from the US SODA model~\cite{carton2005soda}. Each sample encapsulates meteorological and spatial variables, such as sea surface temperature anomalies, heat content anomalies (T300), latitudinal wind anomalies, and longitudinal wind anomalies, with data dimensions (year,month,lat,lon). The training data offers Nino3.4 index-labeled data for the corresponding month. The testing data comprises 12 randomly selected time series from multiple international oceanographic data assimilation results.

\textbf{Extreme-ERA5}~\cite{nguyen2023climatelearn} is a subset constructed by ClimateLearn from ERA5 to evaluate the prediction capability of data-driven models under extreme weather conditions. It comprises various extreme weather events, defined by climate variables exceeding localized thresholds (e.g., heatwaves and cold breaks due to sea level temperature anomalies). The dataset covers the period 1979-2018, with 1979-2015 considered the training dataset.

\textbf{PRISM}~\cite{nguyen2023climatelearn} is a dataset contains myriad observed atmospheric variables, including but not limited to temperature and precipitation for the conterminous U.S. region. Maintained by the PRISM Climate Organization at Oregon State University, the dataset spans from 1895 to the present. At its highest resolution, it provides daily data based on 4 km x 4 km grid cells, forming a matrix of shape 621 x 1405.

\textbf{DroughtED}~\cite{minixhofer2021droughted} is a drought forecast data that combines 180 daily weather observations for the continental United States and geospatial location metadata for all 3,108 counties. It includes meteorological real-time and historical data from NASA's Global Energy Resources (Electricity) Prediction Program, and variables include measurements of precipitation, surface pressure, relative humidity dew/frost point, wind speed, and daily resolution temperature. Past drought observations were also included and given a parallel categorization of USDM drought levels, including no drought (none), abnormally dry (D0), moderate (D1), severe (D2), extreme (D3), and abnormal (D4). Additionally considering that drought is a seasonal phenomenon, seasonal characteristics were also included within the dataset. In addition, a location metric was included, including topographic slope, gradient, and elevation for each site, as well as land use (e.g., rain-fed cropland or forested land) for each site, and soil quality, such as toxicity or nutrient utilization.

\textbf{Digital Typhoon}~\cite{kitamoto2023digital} is a image-based dataset utilized to long-term spatio-temporal modeling for tropical cyclones. It is created from the comprehensive satellite image archive of the Japanese geostationary satellites eries, Himawari, from Himawari-1 to Himawari-9. The dataset consists of 1,099 typhoons and 189,364 images. Geographically, it covers the complete record of typhoons occurring in the Northwestern Pacific region, with a time span from 1978 to 2022. The dataset has a temporal resolution of one hour and a spatial resolution of 5km.

\textbf{EarthNet2021}~\cite{requena2021earthnet2021} is large dataset for Earth surface prediction, extreme summer prediction and seasonal cycle prediction. It contains more than 32,000 samples of Sentinel 2 (high temporal and spatial resolution Earth satellite) Class 2A imagery, as well as daily climate data derived from the E-OBS observational dataset containing interpolated ground-truth observations of weather from multiple stations across Europe for the full year 2018.

\textbf{ClimateNet}~\cite{kashinath2021climatenet} is an open and expert-labeled dataset designed for high-precision analyses of extreme weather events. It focuses on capturing tropical cyclones and atmospheric rivers in high-resolution climate model outputs, simulating the recent historical period from 1996 to 2010. This dataset is valuable for various applications in machine learning and climate research, such as transfer learning, curriculum learning, active learning, spatiotemporal segmentation, probabilistic segmentation, and hypothesis testing.

\textbf{IowaRain}~\cite{sit2021iowarain}  is primarily derived from the Quantitative Precipitation Estimation System (QPES) based on the National Weather Service's Weather Detection Radar Network lowa Flood Center. It covers the region of Iowa and spans the period from 2016 through the end of 2019. Each event in the dataset includes a collection of 2D rainfall rate maps, along with information about the size of the event (i.e., the number of rainfall rate maps in the set) and the start date of the event. This dataset is specifically designed for predicting regional rainfall events.

\textbf{ExtremeWeather}~\cite{racah2017extremeweather} is a comprehensive dataset that aims to facilitate the detection, localization, and understanding of extreme weather events. It is based on post-processed simulations of CAM5, a widely used atmospheric 3D model for global climate simulations. The dataset focuses on extreme weather events and provides a spatial resolution of 25-km. Each snapshot of the global atmospheric state is represented as a 768 × 1152 grid, with 16 simulated climate variables including surface temperature, surface pressure, precipitation, latitudinal winds, meridional winds, humidity, cloud fraction, and water vapor. The dataset covers a time span from 1979 to 2005, with a temporal resolution of 3 hours. It consists of a total of 78,840 samples, capturing four types of extreme weather events: Tropical Depression (TD), Tropical Cyclone (TC), Extratropical Cyclone (ETC), and Atmospheric Rivers (AR). The center of each storm is considered as the reference point for marking the bounding box coordinates. Notably, the dataset includes 39,420 labeled images, providing valuable annotations for training and analysis purposes.

\textbf{KoMet}~\cite{kim2022benchmark} is a collection of data specifically gathered in Korea. It utilizes input data from GDAPS-KIM, a global numerical weather prediction model that offers hourly forecasts for various atmospheric variables. The dataset focuses on precipitation prediction and has a spatial resolution of 12 × 12 kilometers, resulting in a spatial size of 65 × 50. The dataset includes two types of variables: pressure level variables and surface variables. These variables provide valuable information for predicting and understanding precipitation patterns in Korea. In terms of the distribution of samples, approximately 87.24\% of the samples in the dataset are classified as "no rain," indicating instances where precipitation is not observed. Around 11.57\% of the samples correspond to rainy conditions, while 1.19\% represent extreme rainfall events.

\textbf{Germany}~\cite{paulat2008gridded} is a precipitation forecasting dataset collected in West Germany. It spans the period from 2011 to 2018 and focuses on precipitation forecasting. The input data for this dataset are derived from the COSMO-DE-EPS forecast, which provides 143 variables representing different atmospheric states. The dataset has a spatial resolution of 36 × 36 for the input data, indicating the grid size used to represent the atmospheric conditions. The output data, representing the precipitation forecasts, have a higher resolution of 72 × 72. In terms of the distribution of samples, approximately 85.10\% of the samples in the dataset are classified as ``no rain`` indicating instances where precipitation is not observed. Around 13.80\% of the samples correspond to rainy conditions, while 1.10\% represent extreme rainfall events.

\textbf{China}~\cite{tang2023postrainbench} is a precipitation forecasting dataset collected in China, provided hourly, 1 km × 1 km resolution, 3-hourly grid-point precipitation data for the rainy season. This dataset lasts from April through October for the 2020 and 2021 seasons. In addition, it includes 3-hour lead time projections from the regional NWP model, including 28 surface and pressure level variables such as 2-meter temperature, 2-meter dew point temperature, 10-meter u and v wind components, and CAPE (Convective Available Potential Energy) values. Each time frame in this dataset covers a sizable spatial region with a grid size of 430 × 815.

\textbf{China-Precipitation/Temperature}~\cite{peng20191} is a high-spatial-resolution monthly precipitation and temperature dataset for China, covering the period from 1901 to 2017. The dataset includes monthly minimum, maximum, and mean temperatures, as well as precipitation data, at a spatial resolution of 0.5 arcminutes (approximately 1 kilometer) for the main land area of China. The dataset was downscaled using the Delta spatial downscaling method from the 30 arcminute Climatic Research Unit (CRU) time series dataset and the WorldClim climatology dataset. It was evaluated using observations collected from 496 weather stations across China during the period from 1951 to 2016. 

\textbf{ClimART}~\cite{cachay2021climart} is a dataset for emulating atmospheric radiative transfer in weather and climate models, with more than 10 million samples from present, pre-industrial. and future climate conditions, based on the Canadian Earth System Model. This dataset of global snapshots of the current atmospheric state from CanESM5 was simulated every 205 hours from 1979 to 2014. CanESM5 has a horizontal grid discretizing longitude to 128 columns of the same size and latitude to 64 columns using a Gaussian grid (8192 = 128 x 64 columns). This resulted in 43 global snapshots per year for the period 1979-2014, totaling over 12 million columns and a raw dataset size of 1.5 TB.

\textbf{MeteoNet}~\cite{larvor2020meteonet} is a multimodal dataset for regional precipitation nowcasting covering a geographical area of 550 × 550 km in the northwestern quarter of France, spanning the years 2016 to 2018. The modalities of the dataset include radar echo observations, earth-observing satellite imagery, ground station observations, weather forecast model data and topographic maps. The ground observation data has a temporal resolution of six minutes and includes meteorological variables such as temperature, humidity, atmospheric pressure, and wind speed measured by 500 ground stations. The radar echoes, on the other hand, are precipitation radar records with a five-minute time resolution, i.e., 12 frames recorded in one hour, including radar reflectivity and rainfall estimates. The satellite data are recorded every 15 min for Cloud Type (CT) and every 1 hour for Channels (visible, infrared). Weather models are also included forecasts from 2 weather models with 2D parameters, generated once a day.

\textbf{RAIN-F}~\cite{choi2021rain} is a pre-processed spatio-temporally aligned multimodal dataset for short-advance rainfall forecasting, which includes radar, ground-based observations, and a variety of summed satellite data, for the time period from 2017 to 2019, with a coverage of the Korean Peninsula. Specifically, nine different atmospheric state variables (one radar, seven ground observations, and one satellite) associated with precipitation variables are included with a temporal resolution of one hour. The ground-based observations include wind direction and speed, humidity, surface pressure, temperature, sea level pressure and precipitation.

\textbf{RAIF-F+}~\cite{choi2021rainfp} is a new version of RAIN-F with new atmospheric variables and TB products, which can also be used to retrieve atmospheric variables from satellite observations or to predict atmospheric state and precipitation, with geographic and temporal coverage identical to that of RAIN-F.

\textbf{ENS-10}~\cite{ashkboos2022ens} is a post-processing dataset for ensemble weather forecasting, consisting of 10 ensemble members spanning 20 years (1998-2017). These ensemble members are generated by perturbing numerical weather simulations to capture the chaotic behavior of the Earth. To represent the three-dimensional state of the atmosphere, ENS10 provides 11 atmospheric variables at 11 different pressure levels and the most relevant variables at the surface, with a resolution of 0.5 degrees. The dataset includes forecast lead times of $T=0, 24, 48$ hours (two data points per week).

\textbf{SEVIR}~\cite{veillette2020sevir} is a collection of temporally and spatially aligned image sequences depicting weather events captured over the contiguous US (CONUS) by GOES-16 satellite and the mosaic of NEXRAD radars. Five different image data types are included, such as the GOES-16 0.6 µm visible satellite channel (vis), 6.9 µm and 10.7 µm infrared channels (ir069, ir107), a radar mosaic of vertically integrated liquid (vil), and total lightning flashes collected by the GOES-16 geostationary lightning mapper (GLM) (lght). The spatial resolution is 0.5km, 2km, 2km, 1km, and 8km, the temporal resolution is 5 minutes (except for lightning events), and the image coverage is 768×768, 192×192, 192×192, and 384×385, respectively, corresponding to meteorological events 1403, 13552, 13541, 20393, and 15115, which can be used by the applied to weather prediction, image-to-image conversion, extreme weather detection, weather annotation, super resolution and other applications.

\textbf{SRAD2018} is a precipitation nowcasting dataset composed of a series of radar echo image, is from \textit{Tianchi IEEE International Conference on Data Mining (ICDM) 2018 Global Artificial Intelligence Challenge on Meteorology} and collected by Shenzhen Meteorological Bureau and Hong Kong Observatory. Each sequence in the dataset contains $501\times501 $ km region with $1\times1$ spatial resolution, the temporal resolution is 6 min and complete sequence is 6 h, taken from an altitude of 3 km.

\textbf{RainBench}~\cite{de2021rainbench} is a precipitation forecasting dataset consists of European Centre for Medium-Range Weather Forecasts simulated satellite data (SimSat), ERA5 reanalysis product and Integrated Multi-satelliteE Rettrievals (IMAGE) global precipitation estimates. All data is converted from their original resolution to 5.625 resolutions using bilinear interpolation. The time span is 2000 to 2017 and the time resolution is 1 hour.

\textbf{KnowAir}~\cite{wang2020pm2} is a weather forecasting dataset based on station observations, which includes 184 meteorological stations in northern China. The dataset covers the time span from 2015 to 2018, with a temporal resolution of three hours. It primarily includes 18 weather features.

\textbf{Weather2K}~\cite{zhu2023weather2k} is a large-scale dataset for weather prediction based on station observation data, which is extracted from 1,866 ground-based meteorological stations throughout China, covering an area of 6 million square kilometers, with 23 features corresponding to each meteorological battle, containing three static variables representing geographic information as well as 20 interacting meteorological variables, and with a temporal coverage of January 1, 2017, to August 31, 2021, with a temporal resolution of one hour.

\textbf{NASA}~\cite{chen2023prompt} is a collection of regional weather forecasting datasets, which consists of three subsets, AvePRE, SurTEMP, SurUPS, spanning from Apr 1, 2012 to Feb 28, 2016, Jan 3, 2019 to May 2, 2022, and Jan 2, 2019 to Jul 29, 2022, respectively, all with one-hour temporal resolution, collected from 88, 525, and 238 stations, respectively.

\textbf{LSDSSIMR}~\cite{bai2023lsdssimr} is a large-scale dust storm database used for extreme weather and sandstorm prediction. The data is sourced from multi-channel and dust label data of the Fengyun-4A (FY-4A) geostationary orbit satellite, as well as Earth system reanalysis data. The dataset covers the time span from March to May each year from 2020 to 2022, with a time resolution of 15 minutes and a spatial resolution of 4 kilometers. Meteorological reanalysis data is incorporated into LSDSSIMR for spatio-temporal prediction methods. Each data file is stored in HDF5 format, and the final LSDSSIMR consists of nearly 5400 HDF5 files.

\textbf{RainNet}~\cite{chen2020rainnet} is a large-scale dataset specifically designed for spatial downscaling of precipitation. It contains data from 85 months or 62,424 hours, resulting in a total of 62,424 pairs of high-resolution and low-resolution precipitation maps. The high-resolution precipitation maps have a size of 624x999, while the low-resolution maps have a size of 208x333. These data encompass various meteorological phenomena and precipitation conditions such as hurricanes and squall lines. The precipitation map pairs in RainNet are stored in HDF5 files, occupying a total of 360GB of disk space. The data is collected from satellites, radars, and rain gauge stations, covering the inherent working characteristics of different meteorological measurement systems.

\textbf{Continental United States Wind Speeds}~\cite{kurinchi-vendhan_2021} is a climate downscaling (super-resolution) dataset, was obtained from the National Renewable Energy Laboratory's (NREL's) Wind Integration National Database (WIND) Toolkit, with a focus on the continental United States. Wind velocity data is comprised of westward (ua) and southward (va) wind components, calculated from wind speeds and directions 100-km from Earth's surface. The WIND Toolkit has a spatial resolution of 2 km x 1 hr spatiotemporal resolution. The dataset contains data sampled at a 4-hourly temporal resolution for the years 2007 to 2013. The sample test dataset contains data sampled at a 4-hourly temporal resolution for 2014. We transform 2D data arrays of wind speed and direction into corresponding ua and va wind speed components. These are chipped into 100x100 patches. Low resolution imagery is obtained by sampling high resolution data at every fifth data point as instructed by NREL's guidelines.

\textbf{Continental United States Solar Irradiance}~\cite{kurinchi-vendhan_2021} is a climate downscaling (super-resolution) dataset, was obtained from the National Renewable Energy Laboratory's (NREL's) National Solar Radiation Database (NSRDB), with a focus on the continental United States. we consider solar irradiance data from the NSRDB in terms of direct normal irradiance (DNI) and diffused horizontal irradiance (DHI) at an approximately 4-km x 1/2-hr spatiotemporal resolution. The solar dataset produced for this work samples data at an hourly temporal resolution from 6 am to 6 pm for the years 2007 to 2013. The test dataset contains datapoints sampled from 2014. A 1D array of data points is provided along with latitude and longitude metadata for each point. We re-arrange this 1D array into a 2D image based on the lat/long metadata. These 2D arrays of DNI and DHI are chipped into 100 x 100 patches. Low resolution imagery is obtained by sampling high resolution data at every fifth data point.

\subsubsection{Weather and Climate Text Data}
This subsection focuses on weather text datasets, which are more thematically oriented towards climate change related policy statements as well as document texts.

\textbf{CLIMATE-FEVER.}~\cite{diggelmann2021climatefever} is a dataset adopting the FEVER methodology that consists of 1,535 real-world claims regarding climate-change. Each claim is accompanied by five manually annotated evidence sentences retrieved from Wikipedia that support, refute or do not give enough information to validate the claim. The total dataset thus contains 7,675 claim-evidence pairs. Furthermore, the dataset features challenging claims that relate multiple facets and disputed cases of claims where both supporting and refuting evidence are present.

\textbf{ClimateBERT-NetZero.}~\cite{schimanski2023climatebertnetzero} is an expert-annotated dataset from the Net Zero Tracker Project that assesses targets for reduction and net zero emissions or similar aims (e.g., zero carbon, climate neutral, or net negative). The dataset contains 273 claims by cities, 1396 claims by companies, 205 claims by countries, and 159 claims by regions.

\textbf{ClimaText.}~\cite{varini2021climatext} is a dataset for climate change topic detection, consists of labeled sentences. The label generated heuristically or via a mannual process. indicates whether a sentence talks about climate change or not. All sentences are collected from Wikipedia, the U.S. Securities and Exchange COMMISSION (SEC) 10K files. For Wikipedia, collect 6,885 documents, 715 relevant to climate change and 6,170 not relevant to climate change.

\textbf{CLIMA-INS}~\cite{laud2023climabench} contains survey from annual NAIC Climate Risk Disclosure Survey responses for the years 2012-2021, the purpose of the survey is to enhance transparency about how insurers manage climate-related risks and opportunities to enable better-informed collaboration on climate-related issues, where each survey consists of eight questions.

\textbf{CLIMA-CDP}~\cite{laud2023climabench} is composed of three subset part where each part is a set of questionnaires filled out by a city, company, or state respectively. The dataset can performs topic classification and question classification. The number of sample from train, development, and test for task of topic classification is 46.8K, 8.7k, and 8.9K, respectively. In addition, the number of sample from train, development, and test for task for question answering task is 48.2K (8.7K for states, 34.5K for corporations), 8.5K (0.9K for states, 34.5K for corporations), and 9.3K (1.1K for states, 4.9K for corporations), respectively. The number of classes for topic classification task is 12, for question answering is 294, 132, and 43 respectively.

\textbf{CLIMATESTANCE \& CLIMATEENG}~\cite{vaid2022towards} is a ternary classification dataset about climate-related text, extracted Twitter data consisting of 3777 tweets posted during the 2019 United Nations Framework Convention on Climate Change. Each tweet was labelled for two tasks: stance detection and categorical classification. For stance detection the authors labelled each tweet as \textit{In Favour, Against or Ambiguous} towards climate change prevention. For categorical classification, the five classes are \textit{Disaster}, \textit{Ocean/Water}, \textit{Agriculture/Forestry}, \textit{Politics}, and \textit{General}.

\textbf{SCIDCC}~\cite{mishra2021neuralnere} is curated by scraping new articles from the Science Daily website~\cite{mishra2021neuralnere}. It contains around 11k news articles with 20 labelled categories relevant to climate change such as \textit{Earthquakes}, \textit{Pollution}, and \textit{Hurricanes}. Each article comprises of a title, a summary, and a body which on average is much longer (500-600 words) than other climate text datasets.

\subsection{Tools and Models}
In this subsection, we collect and compile a rich and usable set of tools and foundation models for modeling weather and climate data.

\begin{itemize}
    \item \textbf{OpenCastKit:} A new global AI weather forecasting project based on FourCastNet and GraphCast. \url{https://github.com/HFAiLab/OpenCastKit}
    \item \textbf{GraphCast:} A foundation model for medium-range global weather forecasting. \url{https://github.com/google-deepmind/graphcast}
    \item \textbf{FourCastNet:} A foundation model for weather and climate data based on AFNO. \url{https://github.com/NVlabs/FourCastNet}
    \item \textbf{PanGu-Weather:} A foundation model for medium-range glocal weather forecasting. \url{https://github.com/198808xc/Pangu-Weather}
    \item \textbf{FuXi:} A forecasting system for 15-day global weather forecast. \url{https://github.com/tpys/FuXi}
    \item \textbf{W-MAE:} A unsupervised learning global weather forecasting model via Masked Autoencoder. \url{https://github.com/Gufrannn/W-MAE}
    \item \textbf{ClimaX:} A versatile climatefoundation model covering forecasting, projection, and downscaling. \url{https://github.com/microsoft/ClimaX}
    \item \textbf{OceanGPT:} A large language model for ocean science tasks trained with KnowLM. \url{https://huggingface.co/zjunlp/OceanGPT-7b}
    \item \textbf{ClimateBert:} An algorithm that enables to analyze climate-risk disclosures along the four main TCFD categories. \url{https://huggingface.co/climatebert}
    \item \textbf{Climate X Quantus:} An XAI toolbox for ML/DL-based climate models. \url{https://github.com/philine-bommer/Climate_X_Quantus}
\end{itemize}

\section{Challenges, Outlook, and Opportunities}
\label{sec:outlook}
The potential pitfalls of AI foundation models in weather and climate (WFMs) data understanding are manifested in a large number of pending challenges to which data-driven models are more susceptible than traditional NWP models. In this section, we identify five main challenge areas and suggest some best practices that should be recognised and implemented in future research, as well as pointing out research opportunities and routes that hold great promise for the future.

\subsection{Post-Processing of Data}
For DL models, the quality of the data is paramount. However, numerous challenges associated with data pose threats to the development of expansive foundation models for weather and climate data understanding, including issues related to data quality and quantity, post-processing costs, scarcity of historical data, non-stationarity, and the underutilization of existing datasets.

\begin{itemize}
    \item \textbf{Data Quality and Quantity.} Large-scale foundation models require comprehensive and high-quality data for robust results. Despite the exponential increase in global climate data~\cite{mukkavilli2023ai}, like ERA5 and CMIP~\cite{rasp2020weatherbench}, general-purpose datasets that are both large-scale and high-quality are seldom available.
    \item  \textbf{Post-processing Costs.} Large models, such as \textsc{Pangu-Weather}~\cite{bi2023accurate} and \textsc{ClimaX}~\cite{nguyen2023climax}, often necessitate costly post-processing for scenario-specific analyses. The analysis of extreme events, for example, presents a unique challenge. These rare events, which are increasingly likely in a non-stationary climate, are often characterized by outliers in climate variables. Their development involves physical processes that span time cycles from weeks to years, complicating the creation of fine-grained annotations~\cite{racah2017extremeweather}.
    \item  \textbf{Underutilization of Existing Datasets.} Large-scale datasets, despite their size, remain underdeveloped due to the enormous post-processing costs. Benchmark datasets like WeatherBench~\cite{rasp2020weatherbench}, WeatherBench2~\cite{rasp2023weatherbench}, OceanBench~\cite{johnson2023oceanbench}, and ClimateLearn~\cite{nguyen2023climatelearn}, which contain post-processed data, are still in early stages of development due to limited data scenarios.
\end{itemize}

The creation of general WFMs hinges upon the availability of rich, large-scale, post-processed datasets. There is substantial scope for deeper analyses and post-processing of these datasets, including understanding anomalous weather events, integrating physical models, and efficient, rational annotations. Overcoming these challenges is key to realizing the full potential of climate foundation models.

\subsection{Development of Multi-Modal Models}
Time series data are often enriched with supplementary information, including textual descriptions. This is particularly beneficial in economics and finance, where forecasting can harness information from textual data sources such as news articles or tweets, in conjunction with digital economic time series data~\cite{jin2023large}.  Analogously, weather and climate analysis can profit from the diverse modalities encompassed in climate data, which include reanalysis data, multimodal observation data (e.g, radar echoes~\cite{chen2022dynamic,Chen_2023}, satellite imagery~\cite{bai2023lsdssimr}, and geographic terrain features~\cite{minixhofer2021droughted}, etc.). The development of models capable of integrating and learning from this rich array of data modalities has the potential to enhance predictive accuracy. However, while efforts have been made to develop weather prediction and meteorological analysis models based on multimodal meteorological data~\cite{choi2021rainfp,veillette2020sevir,minixhofer2021droughted}, these models often exhibit limitations. They are typically confined to specific geographic regions and struggle to accommodate the extensive spectrum of meteorological modes. A salient challenge in constructing multimodal climate foundation models lies in enabling these models to learn joint representations that encapsulate the sequential nature of temporal data and the unique traits of other meteorological modes.

This challenge encompasses understanding and accommodating the disparate temporal and spatial resolutions across modes. For instance, meteorological observations may have an hourly temporal resolution, radar echo data might possess a six-minute temporal resolution and 1-4 km spatial resolution, and satellite images could exhibit half-hourly temporal resolution and a 5-12 km spatial resolution. The task of leveraging information with different temporal and spatial resolutions to construct a robust and powerful climate foundation model is complex. Furthermore, it is a challenge to balance and align multimodal information collected at different time points to achieve more precise fixed-point prediction and analysis. As such, the development of models that can effectively integrate and learn from these diverse data sources remains a challenging but important frontier in the field of weather and climate analysis.

\subsection{Interpretability and Causability}
A significant challenge associated with the use of AI models for weather and climate analysis is the often inscrutable nature of the model's decision-making process. Many DL algorithms are inherently complex and opaque, rendering their decision-making processes unintelligible to users~\cite{wang2023interpretable,liu2023explainable}. For applications such as machine translation and text generation, the interpretability may not be a key concern. In these contexts, it is typically sufficient for the model to display competent performance to meet most requirements. However, in weather and cliamte applications, the interpretability of the model is of paramount importance.

Non-transparent, black-box models can precipitate catastrophic errors in predictions, which could have devastating impacts on society and the environment. To mitigate this interpretability challenge, tools rooted in the concept of Explainable AI (XAI) have been proposed, such as XAITools\footnote{\url{https://github.com/IntelAI/intel-xai-tools}}, InterpretML\footnote{\url{https://github.com/interpretml/interpret}}, SHAP\footnote{\url{https://github.com/shap/shap}},  LIME\footnote{\url{https://github.com/marcotcr/lime}}, and AI Explainability 360\footnote{\url{https://github.com/Trusted-AI/AIX360}}, etc. These tools aim to bring increased transparency and trustworthiness to black-box models, including those used in various fields such as Earth sciences~\cite{bommer2023finding} (Climate X Quantus\footnote{\url{https://github.com/philine-bommer/Climate_X_Quantus}}), and offer new insights for refining models that underperform. However, these interpretability tools are not without their shortcomings and can exhibit significant biases. In some cases, the truthful representation of the model may depend more on the specifics of the application and its settings, which can render the results difficult to interpret. This suggests that the interpretability insights of climate AI are influenced more by the network's architecture than by the causal inference of weather and climate data.

Unless appropriately designed, Weather/Climate AI may base predictions on non-physical relationships or false correlations. This limitation in drawing causal conclusions from climate models using XAI tools refers to the limited causality these tools can provide. Physics-guided AI, also known as knowledge-guided or physics-informed AI, is one avenue researchers are exploring to impose physical realism and mitigate the effects of false correlations on predictive algorithms~\cite{chen2021physics,feng2023physics,meng2021physics,lutjens2021physically}. However, research in this area is still nascent. Thus, while strides have been made in enhancing the interpretability of AI models, substantial challenges remain, highlighting the need for continued research and development in this critical area.

\subsection{Generalizability of Models}
The generalization capability of a model refers to its competence in making effective predictions beyond the spatiotemporal confines of its training dataset. Lots of DL techniques operate on the assumption of independent and identically distributed (IID) training and test data~\cite{wang2023interpretable}. This implies that the weights calculated during model training remain efficacious even on unseen datasets. However, when applied to weather and climate analysis, foundation models may exhibit suboptimal performance when predicting Non-IID data beyond the training dataset. A notable example of this is the use of foundation models for the prediction of extreme events outside their trained distribution. These biased and anomalous data often induce significant performance degradation in the model. This is especially the case as the warming climate alters the Earth's spatiotemporal distribution. The existing relationships that currently describe the predictive variables and extreme climate events may no longer apply in the future.

Moreover, climate foundation models are typically pre-trained on general data before being fine-tuned on specific task datasets~\cite{nguyen2023climax}. If the fine-tuning data includes adversarial or noisy examples, the process may introduce vulnerabilities. If the temporal data employed for fine-tuning is not meticulously managed, the model may adopt biases or flaws from this data, leading to compromised robustness in practical applications and unreliable outputs. This underscores the imperative for robust generalization. The advent of physics-informed deep learning represents a promising step towards enhancing the robust generalization of climate models. However, the extension of these models beyond their trained distribution remains an area that is not yet fully explored. This highlights the need for continued research into the generalization capabilities of climate models, particularly in light of the rapidly changing climate and the ever-evolving challenges it presents.

\subsection{Privacy, Adversarial Attacks, and Communication}
Weather and climate data are often of high sensitivity, encapsulating a wealth of climate variables, geographical information, and topographical details dispersed across various regions/ countries~\cite{chen2023prompt,chen2023spatial}. In particular, radar and satellite data are highly sensitive. The training of WFMs using such data poses significant challenges from aspects of centralized training, privacy leaks and adversarial attacks.
\begin{itemize}
    \item  \textbf{Centralized Training Issues.} Models typically undergo pre-training with substantial data before being deployed to various downstream tasks~\cite{bi2023accurate, nguyen2023climax, chen2023fengwu, lam2022graphcast, pathak2022fourcastnet}. However, the centralized training strategy can be fraught with problems. Aggregating sensitive data from different regions or countries onto a central server is neither reliable nor practical due to the inherent risks of data leakage and contamination~\cite{chen2023prompt}.
    \item  \textbf{Privacy Leaks and Adversarial Attacks.} During the fine-tuning process, WFMs often memorize specific details from the datasets, which can potentially compromise private data. Contaminated data also pose a risk of deteriorating model performance. Therefore, the adoption of privacy-preserving techniques to prevent privacy leaks and mitigate adversarial attacks is crucial in the training/fine-tuning of WFMs.

\end{itemize}

Recent studies have introduced the use of differential privacy (DP) techniques or federated learning (FL) to train WFMs~\cite{chen2023prompt,chen2023spatial}, effectively lessening the risk of sensitive climate data leakage~\cite{li2020review}. However, these methods are still confronted with communication challenges.
    
\begin{itemize}
    \item \textbf{Communication Overheads in Federated Learning} Federated learning allows different clients to collaboratively train a global model, with each client maintaining a locally replicated model with consistent structures. During the global aggregation stage, each participant uploads their local model parameters to a cloud server for aggregation. This process results in a significant increase in communication overhead between clients and the server due to the large-scale nature of climate models, posing a serious challenge to computational and hardware costs.
\end{itemize}

\subsection{Continuous Learning and On-device Adaption}
The performance of WFMs, despite showing promising results, can be substantially improved through the application of continuous learning and on-device adaption. \textit{Continual learning}~\cite{wang2023comprehensive}, also referred to \textit{lifelong} or \textit{incremental learning}, is the process of updating a model over time as new data emerges. Given the ever-evolving nature of climate and weather patterns due to natural variability and anthropogenic climate change, this approach proves particularly beneficial. It enables models to adapt to these changes, enhancing their predictive accuracy and robustness. \textit{On-device adaptation}~\cite{lee2020learning} involves the customization of a model based on local data at the point of deployment. It has the potential to boost model performance by enabling adjustments to local climate and weather patterns, which may not be comprehensively captured in global training data. Furthermore, on-device adaptation can minimize the requirement for data transmission, thereby enhancing model efficiency and preserving privacy. However, the implementation of continuous learning and on-device adaptation in models poses several challenges. These include ensuring model stability during \textit{Continual learning} and managing the computational and storage constraints of on-device learning:

\begin{itemize}
    \item \textbf{Maintaining Model Stability.} Models undergoing learning can experience a phenomenon known as "catastrophic forgetting," where a model may forget previously learned patterns after being updated with new data. Balancing the maintenance of model stability while still allowing it to learn from new data poses a significant challenge.
    \item \textbf{Managing Computational and Storage Constraints.} The computational power and storage capacity of a device inherently limit on-device machine learning. Deploying and updating large climate models, on devices with limited resources may prove difficult. Techniques for model compression, efficient computation, and selective model updating are essential to make on-device adaptation of models feasible.
\end{itemize} 
Despite these obstacles, Continual learning and on-device adaptation present a promising avenue for enhancing the performance of climate models.

\subsection{Reproducibility}
Reproducibility stands as a cornerstone principle in the realm of scientific research. The capacity to reproduce results using identical data and methodologies not only reinforces the validity of the findings but also propels further research and innovation. Nevertheless, the pursuit of reproducibility in climate foundation models, poses several formidable challenges:
\begin{itemize}
    \item \textbf{Data Availability and Consistency.} climate foundation models frequently utilise extensive datasets, gathered from a plethora of sources over extended periods. The challenge lies in ensuring the availability and consistency of this data for model reproduction. Data may undergo updates or corrections, and access permissions can fluctuate, thereby adding complexity to reproducibility endeavours.
    \item  \textbf{Model Complexity.} climate foundation models often incorporate sophisticated machine learning architectures, intricate pre-processing steps, and advanced training procedures. Reproducing these models necessitates a comprehensive understanding of all these facets. If any segment of the process is inadequately documented or if specific implementation details are proprietary, model reproduction can become an insurmountable task.
    \item  \textbf{Computational Resources.} Climate foundation models typically demand substantial computational resources for training and inference. Reproduction of these models may be prohibitively costly or technically challenging for researchers lacking comparable resources. This disparity can erect barriers to reproducibility and impede the progress of the broader research community.
    \item  \textbf{Non-Determinism in Training.} Several training processes involve elements of randomness, such as random initialization of weights, shuffling of training data, and stochastic optimization methods. These factors can yield slightly divergent models and results, even when employing the same data and model architecture. Ensuring reproducibility amidst such non-determinism can prove challenging.
    \item \textbf{Model Versioning.} As climate foundation models evolve, new model versions are developed. It's crucial to maintain a record of model versions and align them with the specific results they generated for reproducibility. However, this can become complex and arduous to manage, particularly in large collaborative projects.
\end{itemize}
Addressing these challenges necessitates a united effort from the entire research community. This includes the establishment of standards for data management and model documentation, investment in open-source software and infrastructure, and the cultivation of a research culture underscored by transparency and openness. While these issues are complex, resolving them is paramount to the advancement of climate foundation models research and ensuring its benefits are widely disseminated.

\section{Insight for Foundation Model Designing}
\label{sec:design}
This section presents an intricate examination of the design principles that serve as the foundation of current state-of-the-art (SOTA) WFMs. Its intent is to offer an exhaustive guide and insights for the development of resilient, multipurpose climate foundation models. Five perspectives are covered in this discourse: functional design, fusion of multi-source data, data representation, design of network architecture, and strategy for pre-training/fine-tuning.

\subsection{One Fits All}
Establishing foundation models necessitates a judicious selection of tasks, which influences the data employed, the training strategies deployed, the fine-tuning methodologies adopted, and other associated factors. Foundation models are often viewed as a panacea, pre-trained models that are subsequently fine-tuned for various application-specific tasks. Premier climate foundation models, such as FengWu~\cite{chen2023fengwu} and Pangu-Weather~\cite{bi2023accurate}, prioritize systematic modeling of the Earth system, encompassing the prediction of terrestrial and atmospheric climate variables at distinct spatio-temporal scales. Conventionally, these models are trained using data of spatial resolution derived from the widely accepted ERA5 dataset. In contrast, ClimaX~\cite{nguyen2023climax} adopts an alternate approach, pre-training the foundation model at a coarser resolution and later achieving finer spatial resolution predictions, mappings, or down-sampling via fine-tuning. Thus, the primary technical strategy for the development of Weather and WFMs involves pre-training the models with extensive high-resolution data and then fine-tuning them with minimal effort to demonstrate exceptional performance across a range of downstream tasks.

\subsection{Multi-source Data Fusion}
Weather and climate data primarily fall into spatio-temporal series. Our discussion primarily revolves around spatio-temporal sequence tasks, as delineated in Sec.~\ref{sec:dataset}. Due to the variety of data sources, including but not limited to ground stations, remote sensing devices, and simulation-based climate products, the fusion of information from multiple data sources can explicitly benefit the training process of the foundation model and thus lead to improved performance. However, significant modal differences and heterogeneity among data complicate the realization of multi-source fusion operations. We present here insights into this from two main aspects: \textit{Spatio-Temporal Scales}, \textit{Data Modality}.

\begin{itemize}
    \item \textbf{\textit{Spatio-Temporal Scales.}} Practitioners can implement weather and climate models on a global scale by considering data at multiple spatio-temporal scales simultaneously, most commonly under reanalysis data (see Sec.~\ref{sec:dataset}) by fusing high- and low-resolution data to model both fine- and coarse-global features.

    \item \textbf{\textit{Data Modality.}} Weather and climate data's modal mainly focuses on time series\footnote{The definition of time series in this survey covering  univerate/multivariate/spatio-temporal time series, and video streams.} and text. Fusion of multi-source data for foundation model training for weather and climate can be encouraged to capture interrelated knowledge from different scales and data modalities. Examples precipitation nowcasting and the fusion of multiple neutrals at different pressures for robust global forecasting models. Practitioners can explore simultaneous or staged fusion of multimodal weather data to benefit the foundation model. 
\end{itemize}

\subsection{Data Representation and Model Design}
The robust development of WFMs is contingent on effective data interpretation and representation of weather and climate statistics. This process typically involves two stages: initial data representation construction through pre-training, and application of this representational knowledge to downstream tasks via fine-tuning. Unique representation methods are required given that each data point encodes complex contextual information, unlike the features of natural images. The subsequent discourse seeks to address two pivotal questions in this domain: (1) \textit{Which network architectures can effectively represent weather and climate data?} and (2) \textit{What strategies can improve models and facilitate efficient and accurate representations?}

\subsubsection{\textit{Which network architectures can effectively represent weather and climate data?}}

Reanalysis weather and climate datasets bear significant resemblances to natural images, most notably using grid cells to delineate local semantic information. Consequently, almost all network architectures employed in computer vision can be utilised for processing weather and climate grid data, including but not limited to ResNet, U-Net, Vision Transformer, generative adversarial networks (GANs), and diffusion models.

These models serve as the backbone of WFMs designs such as ClimaX~\cite{nguyen2023climax}, PanGu-Weather~\cite{bi2023accurate}, FourCastNet~\cite{pathak2022fourcastnet}, DYffusion~\cite{cachay2023dyffusion},  GraphCast~\cite{lam2022graphcast}. These models strive to establish more efficient relationships between different regions, atmospheric pressure levels, and atmospheric/surface meteorological variables, by leveraging diverse Earth system modeling techniques.

Therefore, when considering the architectural design of foundational models, choices can be made from the foundational architectures of CNNs, RNNs, Transformers, Graph models, GANs, and Diffusion models. When necessary, these models can be combined to enhance the representation capabilities. Detailed descriptions of these models can be found in Sec.~\ref{sec:modelinfo}.

\subsubsection{\textit{What strategies can enhance models and facilitate efficient and accurate representations?}}
Accurate representation of the latent semantic information in weather and climate data hinges on jointly modeling the temporal, spatial, and variable dimensions of the data. Potential strategies to enhance these models include tokenization strategies, positional encoding, attention mechanisms, and time feature extraction. Here we discuss these four strategies in detail:

\begin{itemize}
    \item \textbf{Tokenization Strategy.} The term "token" originated in the context of Transformers, where a critical operation is dividing the original input image into small blocks of local semantic information based on a patch size - a process referred to as tokenization. For irregular reanalysis gridded weather and climate data, the absence of specific rules or definitions for segmentation implies that the choice of tokenization significantly impacts model performance. For instance, ClimaX introduces a coherent tokenization operation~\cite{nguyen2023climax}, while PanGu-Weather~\cite{bi2023accurate}, FuXi~\cite{chen2023fuxi}, and FengWu~\cite{chen2023fengwu} use different methods for encoding variables. A good tokenization strategy should consider sptaio-temporal correlations of different variables while accounting for different physical scales, without introducing excessive complexity.
    \item \textbf{Positional Encoding.} Positional encoding in a Transformer provides spatial information about data points in a sequence. For weather and climate data, different positional encoding strategies can be employed. Compared to fixed encoding, learnable encodings offer more flexibility, as their positional parameters can be updated to increase model robustness.
    \item \textbf{Attention Mechanisms.} Attention mechanisms are critical in Transformers for modeling dependencies between different elements in a sequence. For weather and climate data, attention mechanisms can help capture relationships between different time steps, geographical locations, and meteorological variables. The computational complexity of attention mechanisms also needs to be considered, as many models encounter high cost and reduced speeds during training and inference.
    \item \textbf{Time Feature Extraction.} Weather and climate data contain a temporal dimension, and extracting time features is crucial for model accuracy. Various methods can be employed to extract time features, which can then be used as part of the model input to aid in better understanding and modeling temporal correlations.
\end{itemize}

\subsection{Learning Strategies}
Pre-training a WFM on large-scale datasets not only hinges on costly data post-processing, but also a substantial investment in computational resources. A prime example of this is PanGu-Weather, whose pre-training necessitates over 60TB of high-resolution data and more than 3000 GPU-days on V100-80G~\cite{bi2023accurate}, underscoring the immense scale of the model. In this section, we primarily delve into a variety of learning strategies. These strategies are designed to alleviate the computational burden, thereby facilitating the training and fine-tuning of foundational models for weather and climate tasks.
\begin{itemize}
    \item \textbf{Self-Supervised Learning.} Self-supervised Learning (SSL) is an unsupervised paradigm wherein models are assigned the task of predicting certain components of their own input data. This approach generates labels intrinsically from the data, obviating the need for external annotations. As a result, SSL can exploit copious amounts of unlabeled data for training. Within the sphere of weather and climate modeling, SSL could be utilized to identify climate patterns and trends. For instance, future meteorological conditions could be forecasted using historical weather variables such as temperature, humidity, and wind velocity. This could be accomplished by projecting the subsequent data point within a pre-established temporal window. In so doing, the model can apprehend inherent weather data trends and patterns. The principal advantage of SSL lies in its capacity to harness vast quantities of unlabeled data for training. Furthermore, it can reveal inherent data patterns and structures, which is especially advantageous for weather and climate tasks~\cite{man2023w}.
    \item  \textbf{Semi-Supervised Learning.} Semi-supervised Learning (SML) represents an intermediate approach between Fully Supervised Learning (FLSL) and Self-Supervised Learning (SSL), leveraging both labeled and unlabeled data for model training. This method is particularly advantageous for weather and climate prediction tasks due to the potential scarcity of labeled weather data and abundance of unlabeled data. One prevalent methodology in SML is self-training. Initially, a supervised model is trained using the available labeled data. Subsequently, this model is applied to predict labels for the unlabeled data, which are then employed as pseudo-labels for retraining the model. This iterative process continues until the model's performance plateaus. The salient advantage of SML is its capacity to concurrently utilize labeled and unlabeled data for training. This facilitates an enhancement in model performance, especially when labeled data is limited, by capitalizing on the extensive quantity of unlabeled data.
    \item \textbf{Federated Learning.} Federated learning~\cite{mcmahan2017communication} (FL) is a distributed ML paradigm with the central goal of enabling multiple participants to collaboratively train a model, all while safeguarding data privacy and security~\cite{chen2023collaborative}. In FL, every participant trains their model locally and shares only model updates, rather than the raw data. This endows FL with a distinct advantage when dealing with sensitive data, while also permitting cross-learning from diverse data sources that might be geographically dispersed or unable to be centralized due to privacy or other reasons. In the context of training WFMs, the application of federated learning carries significant benefits. Firstly, meteorological bureaus and research institutions across the globe possess extensive climate data, but owing to data ownership, privacy, and security concerns, this data cannot easily be centralized for processing. Federated learning enables these institutions to collaboratively train a robust weather forecasting model without the direct sharing of data. Secondly, given the typically large scale of weather and climate data, data transfer could potentially become a bottleneck. With FL, data can be processed and trained locally, requiring only the transfer of model updates, thus significantly reducing data transmission demands. Lastly, FL allows the model to benefit from climate data from different geographical locations and types, enhancing the model's generalization ability and accuracy. Currently, numerous studies have incorporated FL into the process of training WFM~\cite{chen2023prompt,chen2023spatial}.
\end{itemize}

\section{Conclusion}
\label{sec:conclusion}
In conclusion, we present a comprehensive and up-to-date survey of data-driven models tailored to analyze weather and climate data. The intention is to offer a fresh viewpoint on this evolving discipline through a systematically organized appraisal of pertinent models. We distill the most salient methodologies within each category, investigate their respective advantages and drawbacks, and propose viable trajectories for forthcoming exploration. This survey is intended to act as an impetus to kindle sustained interest and nurture a persistent enthusiasm for research within the realm of data-driven models for weather and climate data understanding.

\ifCLASSOPTIONcaptionsoff
  \newpage
\fi



%



\bibliographystyle{IEEEtran}
\bibliography{ref}{}

\end{document}